\documentclass[pdflatex,sn-mathphys-num]{sn-jnl}

\usepackage{graphicx}%
\usepackage{multirow}%
\usepackage{amsmath,amssymb,amsfonts}%
\usepackage{amsthm}%
\usepackage{mathrsfs}%
\usepackage[title]{appendix}%
\usepackage{xcolor}%
\usepackage{textcomp}%
\usepackage{manyfoot}%
\usepackage{booktabs}%
\usepackage{algorithm}%
\usepackage{algorithmicx}%
\usepackage{algpseudocode}%
\usepackage{listings}%
\usepackage{longtable}
\usepackage{rotating}

\theoremstyle{thmstyleone}%
\theoremstyle{thmstyletwo}%

\theoremstyle{thmstylethree}%

\begin{document}

\title[Article Title]{FloraSyntropy-Net: Scalable Deep Learning with Novel FloraSyntropy Archive for Large-Scale Plant Disease Diagnosis}


\author*[1,3]{\fnm{Saif Ur Rehman} \sur{Khan}}\email{saif\_ur\_rehman.khan@dfki.de}

\author*[1,2]{\fnm{Muhammad Nabeel} \sur{Asim}}\email{muhammad\_nabeel.asim@dfki.de}
\author[1,2]{\fnm{Sebastian} \sur{ Vollmer}}\email{sebastian.vollmer@dfki.de}
\author[1,2,3]{\fnm{Andreas} \sur{Dengel}}\email{andreas.dengel@dfki.de}

\affil[1]{\orgdiv{German Research Center for Artificial Intelligence}, \orgaddress{ \city{Kaiserslautern}, \postcode{67663}, \country{Germany}}}
\affil[2]{\orgdiv{Intelligentx GmbH (intelligentx.com)}, \orgaddress{ \city{Kaiserslautern}, \country{Germany}}}
\affil[3]{\orgdiv{Department of Computer Science}, \orgname{Rhineland-Palatinate Technical University of Kaiserslautern-Landau} \orgaddress{ \city{Kaiserslautern}, \postcode{67663}, \country{Germany}}}


\abstract{
Early diagnosis of plant diseases is critical for global food safety, yet most AI solutions lack the generalization required for real-world agricultural diversity. These models are typically constrained to specific species, failing to perform accurately across the broad spectrum of cultivated plants. To address this gap, we first introduce the FloraSyntropy Archive, a large-scale dataset of 178,922 images across 35 plant species, annotated with 97 distinct disease classes. We establish a benchmark by evaluating numerous existing models on this archive, revealing a significant performance gap. We then propose FloraSyntropy-Net, a novel federated learning framework (FL) that integrates a Memetic Algorithm (MAO) for optimal base model selection (DenseNet201), a novel Deep Block for enhanced feature representation, and a client-cloning strategy for scalable, privacy-preserving training. FloraSyntropy-Net achieves a state-of-the-art accuracy of 96.38\% on the FloraSyntropy benchmark. Crucially, to validate its generalization capability, we test the model on the unrelated multiclass Pest dataset, where it demonstrates exceptional adaptability, achieving 99.84\% accuracy. This work provides not only a valuable new resource but also a robust and highly generalizable framework that advances the field towards practical, large-scale agricultural AI applications.
}




\maketitle
\begin{table}[h!]
\centering
\begin{tabular}{cl}
\hline
\textbf{Abbreviation} & \textbf{Detail form} \\ \hline
FL & federated learning \\ 
DL & deep learning \\ 
SOTA & state-of-the-art \\ 
CNN & Convolution Neural Network \\ 
AI & Artificial intelligence \\ 
MAO & Memetic Algorithm Optimization \\ 
FedAvg & Federated Averaging \\ 
TRIPOD & Transparent Reporting of a multi prediction model for Individual Prognosis or Diagnosis \\ \hline
\end{tabular}
\end{table}
\section{Introduction}\label{sec1}
Plant diseases represent a significant threat to global food security and agricultural sustainability by causing significant losses in crop yields and economic damage worldwide \cite{gai2024} . These diseases disrupt essential plant functions by leading to reduced crop production and quality. Plant diseases are typically classified based on the pathogens responsible including fungi, bacteria, and viruses \cite{vijayreddy2024}. The timely and accurate diagnosis of these diseases is crucial to control their further spread and minimize their adverse impact on crop yields \cite{george2025}. The importance of diagnosing plant diseases arises from their significant impact on agricultural productivity and economic stability. These diseases result in considerable yield losses, reduced species diversity, and increased costs for mitigation efforts \cite{sambana2025}. Continuous monitoring and time-series analysis can further enhance early detection by facilitating timely intervention before symptoms become severe \cite{laxmi2024}. Symptoms of plant diseases vary depending on the pathogen and plant species, but common signs include water-soaked lesions, necrotic spots, mottling, necrosis, stunting, leaf curling, distortion, vein chlorosis, and ring spots \cite{mduma2015}. These visual cues are often the first signs of infection and are crucial for initial diagnostic assessment. Traditionally, diagnosing plant diseases has relied on manual inspection, serological tests and visual assessment by trained experts, which often lack both in precision and scalability \cite{mduma2015,pai2025}. These methods typically involve observing symptoms that are followed by microscopic examination of plant tissues or pathogen cultures. While these approaches have been widely used but have notable limitations, such as being labor-intensive, health risks, time-consuming, environmental pollution, and often prone to errors \cite{akhtar2024}. Additionally, the lack of knowledge among local farmers, the high costs of consulting experts, and the inherent inaccuracy of manual assessments further complicate disease diagnosis \cite{upadhyay2025}. These challenges emphasize the urgent need for automated, efficient, and accurate methods for detecting and classifying plant diseases.

Artificial intelligence (AI) techniques have emerged as powerful tools for automating plant disease diagnosis \cite{naveed2024}. Deep learning (DL) models, especially convolutional neural networks (CNNs) have demonstrated remarkable capabilities in image recognition and classification by making them highly suitable for analyzing visual symptoms of plant diseases from images of leaves, stems, and fruits \cite{joseph2023,kabir2021}. These models autonomously extract complex features directly from raw image data for enabling accurate and rapid identification of various plant diseases. DL-based systems have shown promising results in differentiating between healthy and multiple diseased plants \cite{kheir2025}. However, these systems require large, diverse, high-quality datasets for training that are challenging to acquire and annotate. Additionally, privacy concerns and high computational demands limit the scalability of centralized DL frameworks, especially when dealing with sensitive agricultural data from multiple sources \cite{gavai2025}. 

The aforementioned limitations of traditional DL including its dependence on aggregated data and the associated privacy risks underscore the urgent need for alternative approaches. FL offers a compelling solution by facilitating collaborative model training across distributed datasets without the exchange of raw data \cite{fahim2024,hari2025}. In the field of plant disease diagnosis, FL enables agricultural farms and research institutions to collaboratively develop robust classification models while maintaining the privacy of sensitive local data \cite{hari2024}. This decentralized framework effectively addresses privacy and governance concerns, while promoting the creation of more generalized and accurate models by integrating diverse datasets from various geographical regions and environmental conditions. Ultimately, the ability of FL to enhance data privacy makes it ideal for advancing secure and scalable AI applications in agricultural disease classification.
\subsection{Study Novelty and Contribution}
The performance of a deep learning (DL) model is a function of its architecture (A), its parameters (Parm), and the dataset (DS) on which it is trained. Formally, the expected performance on a real-world distribution \( P_{DS} \) can be presented as: 

\begin{equation}
\text{Performance}_E = f(A, Parm, DS); \quad \text{where } DS \sim P_{DS}
\tag{EQU-1}
\end{equation}

Literature presents \cite{ref11, ref12} that research in plant disease detection has been constrained by the use of limited, homogeneous, and often small-scale datasets. Let a typical small-scale dataset be defined as \( S_{DS} \), characterized by a limited number of samples \( S_N \).

\begin{equation}
S_{DS} = \{(x_{i}, y_i)\} \quad \text{for } i = 1 \text{ to } S_N
\tag{EQU-2}
\end{equation}

When a model is trained and evaluated solely on such a dataset, it optimizes for a narrow performance metric, often achieving high accuracy \( f(A, Parm, S_{DS}) \). However, this leads to a critical generalization gap. The model performance on the true, global data distribution \( P_{DS} \) is often significantly low:

\begin{equation}
f(A, Parm, S_{DS}) \gg f(A, Parm, P_{DS})
\tag{EQU-3}
\end{equation}

This disparity means that reported high accuracies in prior studies \cite{fahim2024,gavai2025} were likely optimistic and did not translate to reliable performance in diverse, real-world agricultural scenarios. The primary novelty of this work is to directly bridge this gap by constructing a large-scale, globally representative dataset, \( G_{DS} \), and a novel framework, FloraSyntropy-Net, designed to maximize \( f(A, Parm, G_{DS}) \) with the explicit goal of ensuring that this performance is maintained on \( P_{DS} \).

This study makes the following contributions to the field of automated plant disease diagnosis:
\begin{itemize}
    \item \textbf{Introduction of the FloraSyntropy Archive:} We present a large-scale, heterogeneous, and globally representative image dataset for plant disease diagnosis G\_DS, comprising 178,922 images across 97 distinct classes. Curated from 13 public repositories, this dataset directly addresses the critical research gap of small-scale, non-global evaluation and provides a robust benchmark for assessing true model generalizability.
    
    \item \textbf{A Novel Weighted Federated Optimization:} We introduce a robust federated optimization scheme that moves beyond standard FedAvg. Our method incorporates a dynamic weight scaling factor (\( n_k / n \)), which explicitly weights each client contribution to the global model update based on their data volume and class distribution. This ensures the framework is robust to the inherent data heterogeneity and imbalance found in real-world scenarios, leading to more stable and equitable learning.
    
    \item \textbf{Architectural Innovation with the Novel Deep Block:} We design and integrate a new learnable module that enhances the feature extraction capabilities of the base model. This block, which incorporates dense transformation, concatenation, and sequential feature restructuring, was empirically proven to boost overall accuracy by 1.64\% and deliver substantial performance gains on numerous challenging, underperforming classes.
    
    \item \textbf{Demonstration of Exceptional Cross-Domain Generalization:} We rigorously validate the FloraSyntropy-Net framework robustness not only on its primary FloraSyntropy archive but also through cross-domain adaptation on the unrelated multiclass Pest dataset. FloraSyntropy-Net achieved an efficient accuracy of 99.84\%, demonstrating an unparalleled ability to generalize to novel visual features and biological concepts, a capability critically lacking in previous studies.
\end{itemize}
\section{Related work}
DL has become increasingly important in agriculture due to its capacity to process and analyze vast amounts of data \cite{kowalska2023}. This capability offers numerous benefits, such as enhanced productivity, cost reduction, and the promotion of sustainable farming practices. Yakkundimath et al.\cite{yakkundimath2022}  employed DL models based on transfer learning for rice plant disease classification, using pre-trained VGG-16 and GoogleNet CNN models. The models were evaluated on a dataset of 12,000 labeled images representing three rice diseases and 24 symptoms, achieving classification accuracies of 92.24\% and 91.28\%, respectively. Hassan et al.\cite{hassan2022}  proposed a DL model incorporating inception layers, residual connections, and depthwise separable convolution. Their model achieved 76.59\% accuracy on the cassava dataset by demonstrating improved performance with fewer parameters compared to state-of-the-art models. Haque et al. \cite{haque2022} proposed a DL method for identifying diseases in maize crops using a custom dataset collected. Three architectures based on the Inception-v3 framework were trained on this dataset with the top-performing model achieved 95.99\% of accuracy. The robust model was compared with SOTA approaches to assess its performance. DL offers a promising approach for various disease identification through crop images. However, this method raises concerns regarding data privacy, as it often necessitates the sharing of image data from multiple farms. However, the shift towards FL is becoming increasingly relevant due to the growing challenges posed by data heterogeneity. Adaptive federated learning (AFL) \cite{hari2025} methods are gaining significant attention for their potential to enhance model performance in federated settings.
  
  Tripathy et al. \cite{tripathy2024} propose a novel approach using FL for rice leaf disease classification. Their method involves local training on nodes, followed by model aggregation at the server by utilizing pre-processed images, data augmentation, and feature extraction. The LeNet model optimized with the Spotted Hyena Archimedes Optimizer achieved a precision of 91.30\% by showcasing reasonable performance in disease classification. Shrivastava et al.\cite{shrivastava2024} proposed a federated learning-based crop disease classification method using the Vision Transformer model. Their method achieved 92.00\% accuracy in disease detection, which was further improved to 97.00\% with 20 clients, thus ensuring data privacy while enhancing performance. Chorney et al.\cite{chorney2025} proposed a FL-based collaborative classifier for data privacy and efficiency in rice crop disease detection. Their model effectively classified multiple disease types and healthy states by achieving 83.24\% accuracy performance.  The accuracy can be further improved by using more advanced approaches. Hari et al.\cite{hari2024} propose Federated Deep Learning (FDL) for plant leaf disease detection, where local models trained on region-specific datasets share knowledge with a parent model for reducing computational costs. Their method uses the Plant Village dataset and employs a lightweight Hierarchical Convolutional Neural Network (H-CNN) for achieving 93.00\% testing accuracy. Deng et al. [25] propose a FL-based pest detection technique using an improved faster R-CNN with ResNet-101, multisize feature map fusion, and a soft-non maximum suppression (NMS) algorithm. Their method achieves an average detection accuracy of 90.27\% for multiple pests and diseases in apples.
\section{Methodology}
This section details the FloraSyntropy-Net framework, a novel federated learning framework for collaborative model training. We begin by introducing the Novel Dataset FloraSyntropy Archive and the Genetic Algorithm for Global-Net Model Selection used to establish a robust base model. The core of our methodology involves Finetuning the Architecture and Leveraging Federated Collective Knowledge for Client Robustness. We then explain the process of Global FloraSyntropy-Net: Client Cloning and provide a FloraSyntropy-Net Sync: Client Training Guide for implementation. Finally, we present a comprehensive Overview of the framework, followed by its step-by-step Training and Evaluating procedures to validate its performance.
\subsection{Novel Dataset FloraSyntropy Archive}
\begin{figure}[h]
    \centering
    \includegraphics[width = 12cm]{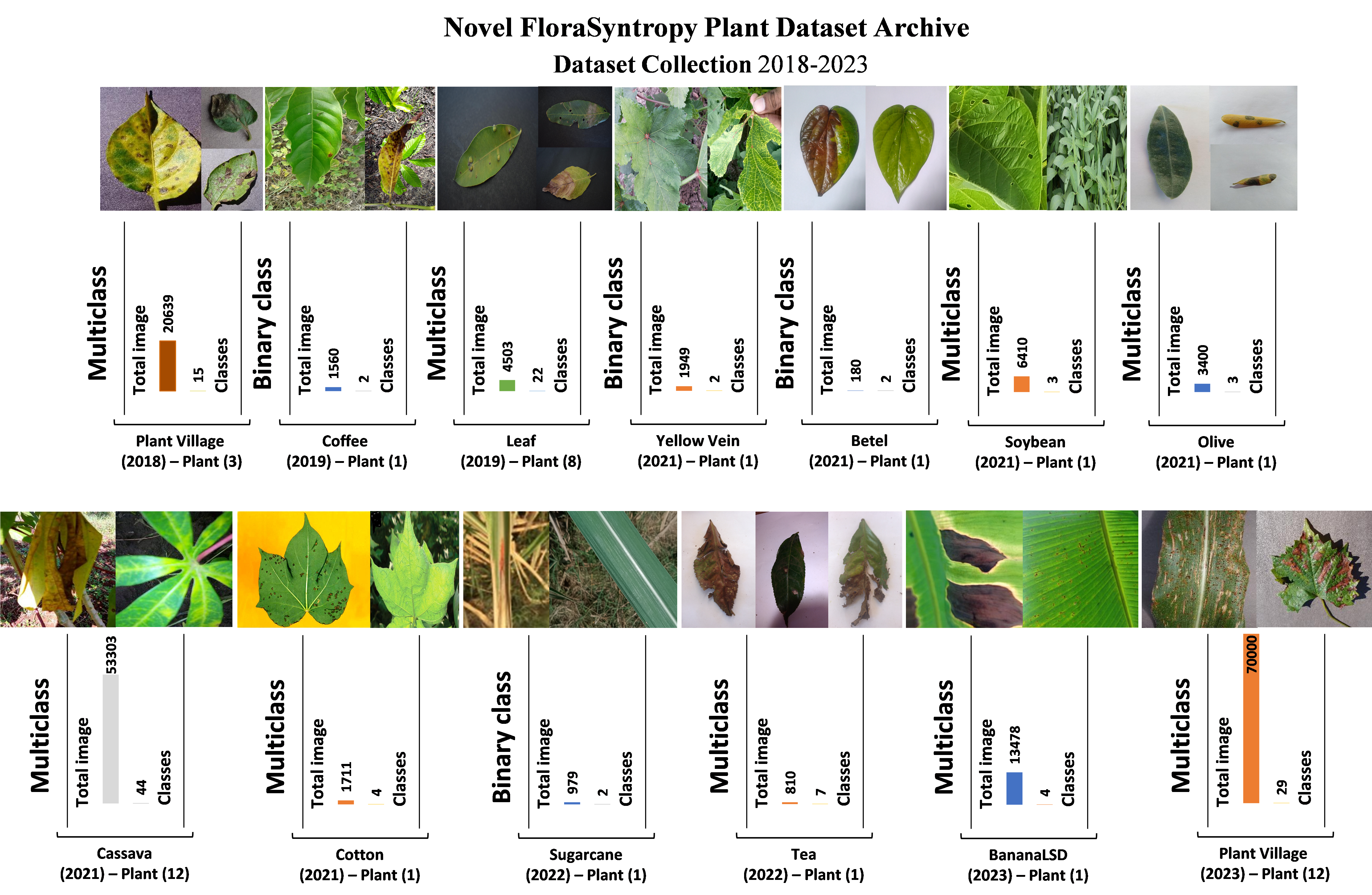}
    \caption{Visual samples from the 13 constituent datasets of the Novel FloraSyntropy Archive, demonstrating the morphological diversity and scale of the collected 178,922 images across 97 classes.}
    \label{fig1}
\end{figure}

A large-scale plant disease dataset is essential for enabling the rapid and accurate diagnosis critical to safeguarding global food security. Such a resource allows for the development of robust AI tools capable of identifying a wide spectrum of threats early. This timely detection is the key to implementing effective, targeted interventions, preventing devastating crop losses, minimizing pesticide overuse, and containing outbreaks before they cause widespread damage. From 2018 to 2023, a significant research gap existed, as no previous studies had comprehensively evaluated models on a large-scale, globally representative plant disease dataset. This limitation meant that the reported high accuracy of many models was often achieved on limited, controlled, or homogenous data, making their true real-world performance and generalizability unclear and potentially overstated.  To directly address the limitation of small-scale, non-global evaluations, this study introduces the Novel FloraSyntropy Archive, a large-scale, heterogeneous dataset compiled from 13 publicly available repositories spanning 2018 to 2023. This archive is meticulously curated to ensure global representation and diversity, culminating in a vast collection of 178,922 images across 97 distinct classes of diseased and healthy plants. The dataset integrates historically significant benchmarks like Plant Village (2018, 20,639 samples) \cite{hlaing2018} with extensive newer additions such as Cassava (2021, 53,303 samples) \cite{fathima2023} and Plant Village V2 (2023, 70,000 samples) \cite{parez2023}, while also including crucial specialized datasets for crops like Banana (BananaLSD, 2023) \cite{shetty2024}, Coffee (2019) \cite{acidri2020}, Soybean (2021) \cite{yoosefzadeh2021}, Tea (2022) \cite{xu2022}, and Sugarcane (2022) \cite{kai2022}. This comprehensive aggregation provides an unprecedented benchmark for developing and fairly evaluating robust, generalizable models for global plant disease diagnosis. Fig \ref{fig1} presents a visual sample from each of the 13 repositories comprising the FloraSyntropy Archive, illustrating the diversity of its 178,242 total images.
\subsection{Memetic Algorithm for Global-Net Model Selection}
Selecting an optimal base model (Global-Net), is a critical step that significantly influences the performance of the entire FloraSyntropy-Net framework. To automate this selection process and identify the most robust and suitable architecture from a vast search space, we employed a Memetic algorithm optimization (MAO) \cite{neri2012}. The MAO is a hybrid metaheuristic that effectively combines the global exploration capabilities of a genetic algorithm with the local refinement efficiency of a local search strategy, making it exceptionally suited for complex optimization problems.
\section*{Utilization and Initialization}

The search process was initialized with a diverse population \( MP_0 \) of \( Pr = 11 \) highly-performing, pre-trained models. This initial gene pool was carefully curated to ensure genetic diversity:

\[
MP_0 = \{ mp_1, mp_2, mp_3, \dots, mp_{11} \} \tag{EQU-4}
\]

where each model \( mp_i \) represents a chromosome defined by its architectural hyperparameters \( hp_i^{arch} \) and its pre-trained weights \( W_i \) from ImageNet. The objective was to find the model that maximizes the validation accuracy after a standardized preprocessing \( TD_{train} \), where \( F \) is the fitness function.

\[
mp^* = \arg\max_{i} F(mp_i) \tag{EQU-5}
\]

\section*{Working Principle of MAO Approach}

The Memetic Algorithm iteratively improves a population of candidate solutions over \( G \) generations. Each generation \( t \) consists of the following steps:

\subsection*{Fitness Evaluation}

Each individual model \( mp_i \) in the current population \( P_i \) is evaluated using a fitness function \( F \), defined as the validation accuracy after a rapid fine-tuning cycle. Let \( D_{val} \) be the validation dataset with \( N_{val} \) samples. For a given model \( mp_i \), let \( \hat{c}_j \) be the predicted class for the \( j \)-th validation sample, and \( C_j \) be the true class. The fitness \( F(mp_i) \) is calculated as:

\[
F(mp_i) = \frac{1}{D_{val}} \sum_{j=1}^{D_{val}} I(\hat{c}_j = C_j) \tag{EQU-6}
\]

where \( I(\cdot) \) is the indicator function that returns 1 if the condition is true and 0 otherwise.

\subsection*{Selection (Tournament Selection)}

To select a parent, a random subset \( T \) of size \( k \) is chosen from the population \( J_t \). The individual in \( T \) with the highest fitness is selected as a parent:

\[
\text{Parent} = \arg\max_{m \in T} F(mp) \tag{EQU-7}
\]

This process is repeated to select a second parent.

\subsection*{Crossover (Recombination)}

Two parents \( Pr_1 \) and \( Pr_2 \) produce an offspring \( \Phi \) by exchanging architectural components. Let the architectural configuration of a model be represented as a vector \( hp \). The offspring architecture \( hp_{\Phi} \) is derived as:

\[
hp_{\Phi} = M \Theta hp_1 + (1 - M) \Theta hp_2 \tag{EQU-8}
\]

where \( M \) is a binary mask vector (determined by uniform crossover), and \( \Theta \) denotes element-wise multiplication.

\subsection*{Mutation}

The offspring’s architectural parameters are mutated with probability \( p_m \). For each parameter \( hp_{\Phi}[i] \) in \( hp_0 \):

\[
hp'_{\Phi}[i] = 
\begin{cases}
hp_{\Phi}[i] + \epsilon & \text{with probability } p_m \\
hp_{\Phi}[i] & \text{otherwise}
\end{cases} \tag{EQU-9}
\]

where \( \epsilon \) is a small random value sampled from a normal distribution \( H(0, \lambda) \).

The algorithm terminates after \( G \) generations, and the model with the highest fitness in the final population \( P_G \) is selected as the Global-Net:

\[
\text{GlobalNet} = \arg\max_{m \in P_G} F(mp) \tag{EQU-10}
\]

which was conclusively identified as DenseNet201 for our framework.
\subsection{Finetune Architecture: FloraSyntropy-Net framework}
The baseline model (Global-Net) for Large-Scale Plant Disease Diagnosis is DenseNet201, a powerful CNN from the DenseNet family \cite{huang2017}. Its core innovation is the Dense Block, where each layer receives the feature maps of all preceding layers as input, connected via concatenation. This promotes feature reuse, strengthens gradient flow, and mitigates the vanishing gradient problem in deep networks. DenseNet201 is composed of an initial convolutional and pooling layer, followed by four sequential dense blocks separated by transition layers (which use 1x1 convolutions for compression and average pooling for downsampling). The final component is a global average pooling layer and a classifier, typically a softmax-activated fully connected layer. Figure 2 presents the architecture diagram of selected (Global-Net) model.
    
\begin{figure}[h]
    \centering
    \includegraphics[width = 11cm]{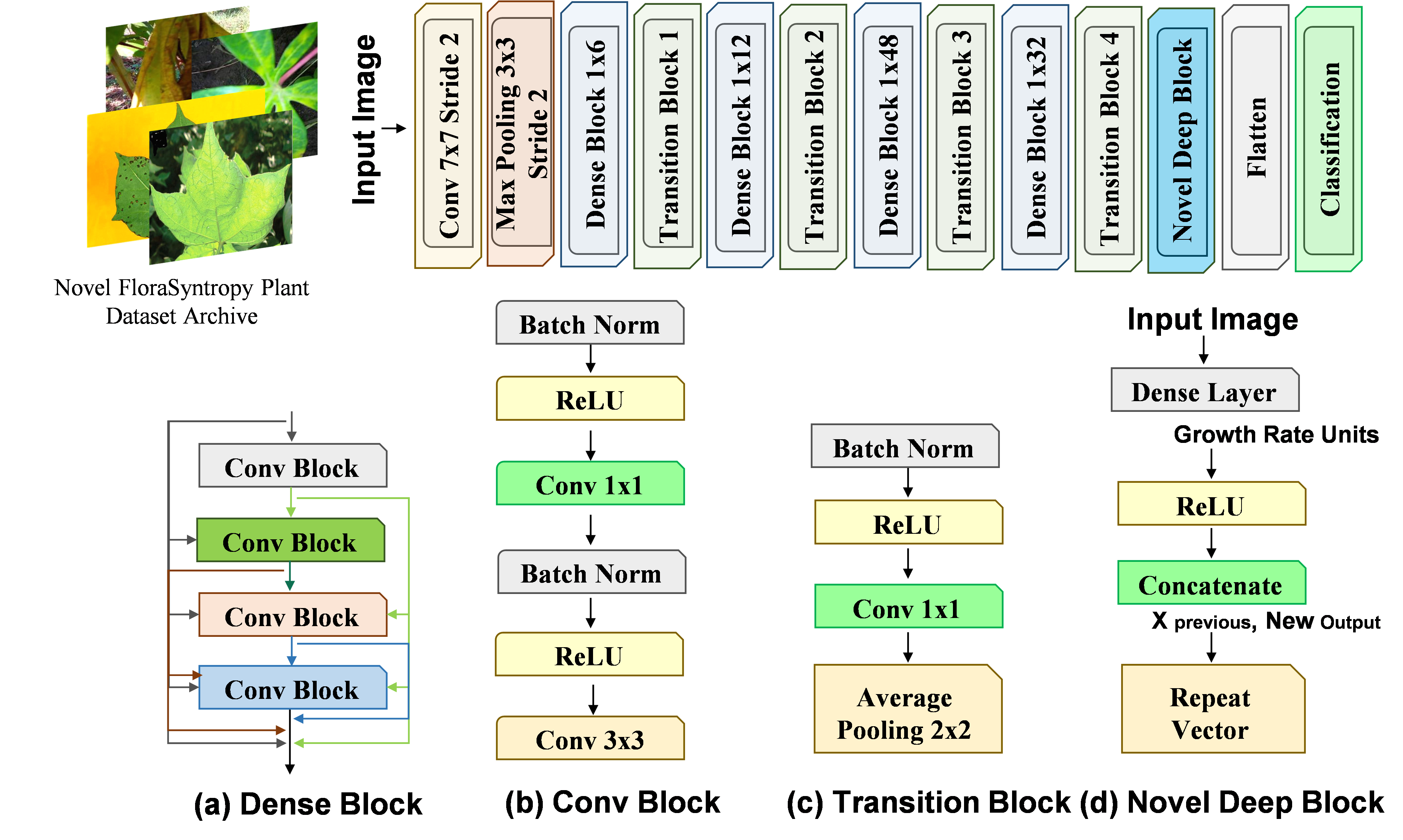}
    \caption{Overview of the Global-Net (DenseNet201) Architecture and Novel Deep Block Integration Mechanism}
    \label{fig2}
\end{figure}

    The finetuning process involves integrating a novel block, termed the Deep Block, after the feature-extracting layers of the DenseNet201 (i.e., before the flatten layer). As depicted in the fig \ref{fig2} diagram, this block is designed to add depth and non-linear representation power to the extracted feature vector before the final classification. The block begins by taking the input vector and passing it through a Dense Layer. The output of this layer is then processed by a ReLU activation function to introduce non-linearity. A key feature of this block is the use of a Concatenate operation, which merges the original input vector with the newly transformed ReLU-activated output. This creates a wider feature vector that preserves original information while incorporating new, higher-level features.
    
    This concatenated output is then fed into a Repeat Vector layer. This layer is crucial as it transforms the 1D feature vector into a 2D sequence by repeating it a fixed number of times, effectively creating a temporal and sequential structure out of the static features. This reshaped data can then be fed into subsequent layers, such as additional 1D layers, for further processing. Finally, the entire Deep Block structure is wrapped in a recursive loop, indicated by the Repeat Vector acting as a loop boundary, meaning this process of dense transformation, concatenation, and repetition is performed multiple times to progressively build deeper and more complex feature representations tailored for the specific finetuning task.
\subsection{Leveraging Federated Collective Knowledge for Client Robustness}
Federated knowledge \cite{khan2025} provides a foundational framework for advancing collaborative model training in distributed environments. This paradigm is essential for improving client skills, especially when it comes to multiclass plant disease detection. Models can be trained on a variety of datasets from various customers by utilizing dispersed data sources while maintaining anonymity, which enhances generalization and performance. By this approach, our FloraSyntropy-Net can gain from the federated sharing of collective knowledge from five sequence clients learning, which will ultimately improve diagnosis accuracy across different global plant classes. 

In order to guarantee that each client contribution to the model update is proportionate to the quantity of data clients possess, the weight scaling factor in a Federated Learning (FL) arrangement is crucial. This scaling approach enables the equitable aggregation of model updates from multiple clients, each with varying amounts of training data, in the context of multiclass plant diagnosis. The amount of data points possessed by a particular client \( \text{LocalNet}_\text{Count} \) and the overall number of data points across all clients \( \text{GlobalNet}_\text{Count} \) are the first steps in calculating the weight scaling factor in this study. 

The weight scaling factor, which is determined by the ratio of these two variables, guarantees that clients with more data have a correspondingly greater impact on the global model update. By keeping clients with different quantities of data balanced, this method improves the model's capacity to generalize across various datasets for better disease diagnosis results. The following is a mathematical representation of the weight scaling factor:

\[
\text{Scaling Weight (Factor)} = \frac{\text{C}_\text{LocalNet}}{\text{C}_\text{GlobalNet}} \quad \text{(EQU-11)}
\]

Here, \( \text{C}_\text{LocalNet} \) is a client's total number of data points, as determined by:

\[
\text{C}_\text{LocalNet} = \text{cardi}(T_\text{client}) \times \text{batch size} \quad \text{(EQU-12)}
\]

In this case, \text{batch size} is the number of data points in each batch, and \( \text{cardi}(T_\text{client}) \) is the total number of batches in the client dataset. 

Where, the total amount of data points across all clients \( \text{C}_\text{GlobalNet} \) is determined by:

\[
\text{C}_\text{GlobalNet} = \sum_{k=1}^{n} \left( \text{cardi}(T_\text{client}) \times \text{batch size} \right) \quad \text{(EQU-13)}
\]

The \( T_\text{client} \) is the dataset for the \( k \)-th client and \( n \) is the total number of clients (in our case, it's five). Therefore, each client model update is scaled based on the percentage of data they contribute to the global dataset using the weight scaling factor.
\subsection{Global FloraSyntropy-Net: Client Cloning}
For multiclass global plant disease diagnosis, cloning the Global FloraSyntropy-Net model offers a scalable and computationally effective method. This method ensures low computational overhead while preserving the integrity of the global model by using the same FloraSyntropy-Net model both globally and by each individual client. This approach eliminates the need for intricate, resource-intensive architectures by replicating the global model at the client level, allowing each client to conduct local training. The system strikes a balance between scalability, efficiency, and high-performance multiclass classification for plant diagnosis by utilizing a common model at the client and global levels.  It is efficient for resource-constrained applications since it uses a robust model for both global and client-level tasks, which guarantees quicker training times, less memory usage, and less energy consumption. Additionally, this approach reduces the frequency of data transfers between clients and the global model, which lessens the computational load and boosts system effectiveness.

FedAvg: Federated Averaging (FedAvg) equation for the global model is updated after each round of training presents as below:
\begin{itemize}
    \item Let \( w^t \) be the global model weights at communication round \( t \).
    \item Let \( K = 5 \) be the total number of clients.
    \item Let \( S^t \) be a subset of clients selected for training in round \( t \), with \( |S^t| = m \).
    \item Let \( n_k \) be the number of data samples on client \( k \).
    \item Let \( n = \sum_{k=1}^{m} n_k \) be the total number of samples across the selected clients.
    \item Let \( w_k^{(t+1)} \) be the updated weights from client \( k \) after training on its local data.
\end{itemize}

The update of the global model is given by:

\[
w^{(t+1)} = \sum_{k=1}^{m} \frac{n_k}{n} \cdot w_k^{(t+1)} \quad \text{(EQU-14)}
\]

This equation ensures that each client's contribution to the model update is proportionate to the quantity of data they possess. The term \( \frac{n_k}{n} \) is precisely that scaling factor.
\subsection{FloraSyntropy-Net Sync: A Client Training Guide}
A crucial step in FL is client training and global synchronization, especially for global applications like multiclass plant disease detection. Individual clients (such as agriculture planting and research institutes) use their own datasets, which may vary (Local \& Global geographical location) in terms of data and plant disease kinds, to train local models in this framework. The learn parameters are then aggregated while preserving data privacy by regularly synchronizing these local models with a global model. There are multiple steps in the process:
\section*{Individual Training (Clients):}

Every trainer (client) uses its own data to train a local model. This model is FloraSyntropy-Net to balance efficiency and performance in the context of plant disease diagnosis. Only model update gradients are transmitted during client-side training, guaranteeing the privacy of sensitive local client (agriculture planting and research institutes) data. Moreover, every trainer (client) \( \text{Tr}_i \) uses its local dataset \( \text{PlantDB}_i \) to train a FloraSyntropy-Net \( \text{FSyn}_i \). Here, the training goal is to minimize a loss function \( \rho(\text{FSyn}_i; \text{PlantDB}_i) \), and the model is a FloraSyntropy-Net. Following training on client trainer (client) \( \text{Tr}_i \)'s data, the local model parameters \( \hat{\text{FSyn}}_i \) are updated as follows:

\[
\text{FSyn}_i^{\text{new}} = \text{Fyn}_i - \eta \nabla \rho(\text{Fyn}_i; \text{PlantDB}_i) \quad \text{(EQU-15)}
\]

Where:
\begin{itemize}
    \item \( \text{Fyn}_i \) is the client model parameters for trainer (client) \( \text{Tr}_i \).
    \item \( \eta \) is the rate of learning.
    \item \( \nabla \rho(\text{Fyn}_i; \text{PlantDB}_i) \) is the gradient of the loss function with respect to the model parameters \( \text{Fyn}_i \) using data from \( \text{PlantDB}_i \).
\end{itemize}

\section*{Global-Net Synchronization:}

All trainers (clients) \( \text{Tr}_i \) send their model modifications to a central server following local training. The server merges these updates in a process known as global aggregation, typically by averaging (FedAvg) the model parameters. This improves the global model \( \text{FSyn}_i^{\text{new}} \) capacity to generalize across many data sources by enabling it to learn from the combined knowledge of all trainers (clients) \( \text{Tr}_i \). The weighted average of the local models \( \text{FSyn}_\text{local} \) from each trainer (client) \( \text{Tr}_i \) is used to update the global model parameters \( \text{FSyn}_\text{global} \) following local training. The representation of the global model is:

\[
\text{FSyn}_\text{global}^{\text{new}} = \frac{1}{G} \sum_{x=1}^n \text{gradient}_i (\text{FSyn}_i^{\text{new}}) \quad \text{(EQU-16)}
\]

Where:
\begin{itemize}
    \item \( G \) is the total number of trainers (clients) \( \text{Tr}_i \). In our case, it is five trainers (clients) \( \text{Tr}_5 \).
    \item \( \text{gradient}_i = \frac{|\text{Data}_x|}{\sum_{x=1}^5 |\text{Data}_x|} \) is the weight allocated to trainer (client) \( \text{Tr}_5 \), which is proportional to the number of data points \( |\text{Data}_x| \) the trainer (client) \( \text{Tr}_5 \) has.
    \item \( \text{FSyn}_i^{\text{new}} \) are each trainer (client) updated settings.
    \item \( \text{FSyn}_\text{global}^{\text{new}} \) is the post-aggregation global model parameters.
\end{itemize}

While preserving anonymity, our averaging procedure guarantees that the global model benefits from the combined knowledge of all the clients.

\section*{Scalable Learning}

The procedure needs to be effective for the model to scale across several trainers (clients) \( \text{Tr}_i \) with global plants having different volumes of data. The weight scaling factor equation is as follows:

\[
\text{Scaling Weight (Factor)} = \frac{|\text{Data}_x|}{\sum_{i=1}^5 |\text{Data}_x|} \quad \text{(EQU-17)}
\]

Where:
\begin{itemize}
    \item The number of data points in the trainer (client) \( \text{Tr}_i \) dataset is represented by \( |\text{Data}_x| \).
    \item \( \sum_{i=1}^5 |\text{Data}_x| \) is the sum of the data points for every client.
\end{itemize}

Trainers (clients) \( \text{Tr}_i \) with more data are guaranteed to contribute more to the global model thanks to this weight scaling.

\subsection{Overview of proposed FloraSyntropy-Net framework}
\begin{figure}[h]
    \centering
    \includegraphics[width = 12cm]{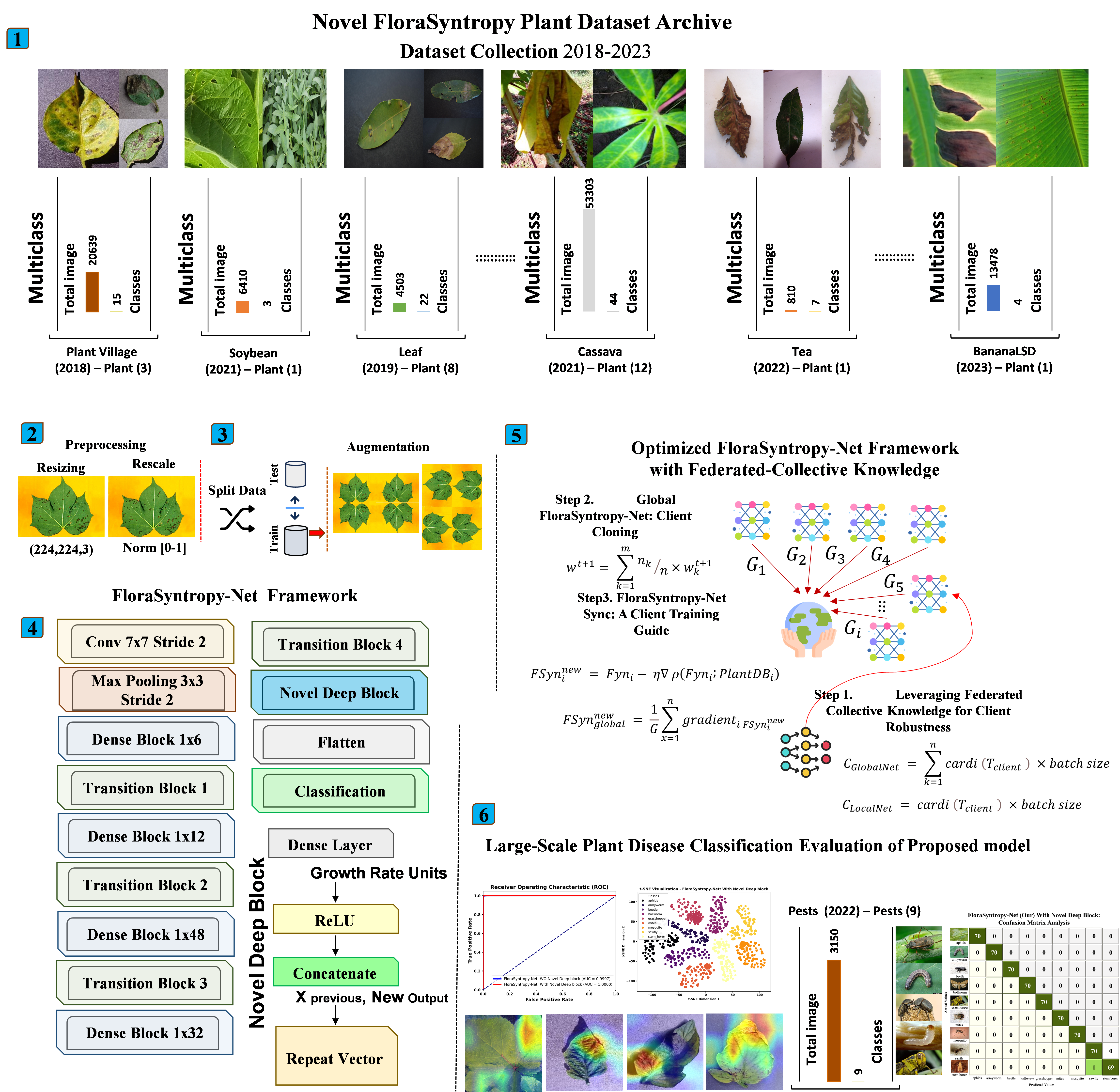}
    \caption{Overview of the proposed FloraSyntropy-Net framework}
    \label{fig3}
\end{figure}

The FloraSyntropy-Net framework represents a comprehensive FL architecture specifically designed to address the challenges of large-scale plant disease diagnosis across heterogeneous, globally-sourced data. The framework begins with a rigorously curated FloraSyntropy Archive, a novel dataset aggregating and standardizing images from 13 public repositories to ensure diversity and representativeness. A critical preprocessing phase resizes all images to 224×224 pixels to maintain dimensional consistency, while advanced augmentation techniques including geometric transformations (balance dataset) are applied to enhance data variety and mitigate class imbalance. This foundational step ensures robust feature extraction and minimizes bias, enabling the model to generalize effectively across varied agricultural contexts. Architecturally, FloraSyntropy-Net incorporates a hybrid optimization approach to maximize performance and adaptability. A MAO is employed for automated base model selection, combining global genetic search with local refinement to identify DenseNet201 as the optimal backbone (Global-Net) for the framework. This base architecture is enhanced with a novel Deep Block, a novel module inserted before the classification layer to augment feature representation. The Deep Block employs dense layers, concatenation operations, and recursive feature restructuring to capture richer spatial hierarchies and nuanced discriminative patterns, significantly improving the model ability to differentiate between visually similar disease phenotypes. For decentralized and privacy-aware training, FloraSyntropy-Net implements a FL structure supported by a weighted aggregation strategy. The \texttt{Global-Net} model is cloned across five multiple clients, each training locally on private data. Model updates are aggregated using a modified Federated Averaging (FedAvg) protocol, where contributions are weighted by client dataset size:
\[
w^{(t+1)} = \sum_{k=1}^{m} \frac{n_k}{n} w_k^{(t+1)}
\]

to ensure equitable influence and robustness against data heterogeneity. This design not only facilitates collaborative learning without data sharing but also enhances the framework scalability and applicability to real-world agricultural networks, where data privacy and computational efficiency are paramount. Fig \ref{fig3} presents the Overview of the proposed FloraSyntropy-Net framework.
\subsection{Training FloraSyntropy-Net framework procedure}
To guarantee effectiveness and consistency, we used a meticulous chosen set of hyperparameters when training the FloraSyntropy-Net. In particular, the model was trained across 30 epochs to allow for adequate learning while avoiding overfitting, and a learning rate of 0.001 was selected to maximize convergence. In order to balance model accuracy and computational performance, a batch size of 32 was chosen. Compared to other optimization methods, the Adam optimizer offered quick convergence and better generalization, which was crucial in obtaining improved performance. It was quite successful because it could adjust the learning rate for each parameter separately, which resulted in more steady training and a lower chance of overshooting.  Furthermore, the categorical cross-entropy loss function was utilized, which was crucial for the classification of plant diseases on a broad scale. This loss function effectively guided the optimization process and ensured a proper evaluation of the model accuracy. Overall, the FloraSyntropy-Net performance significantly improved as a result of the combination of these well-chosen hyperparameters and methodologies, demonstrating the effectiveness and resilience of the suggested architecture.
   
\begin{figure}[h]
    \centering
    \includegraphics[width = 11cm]{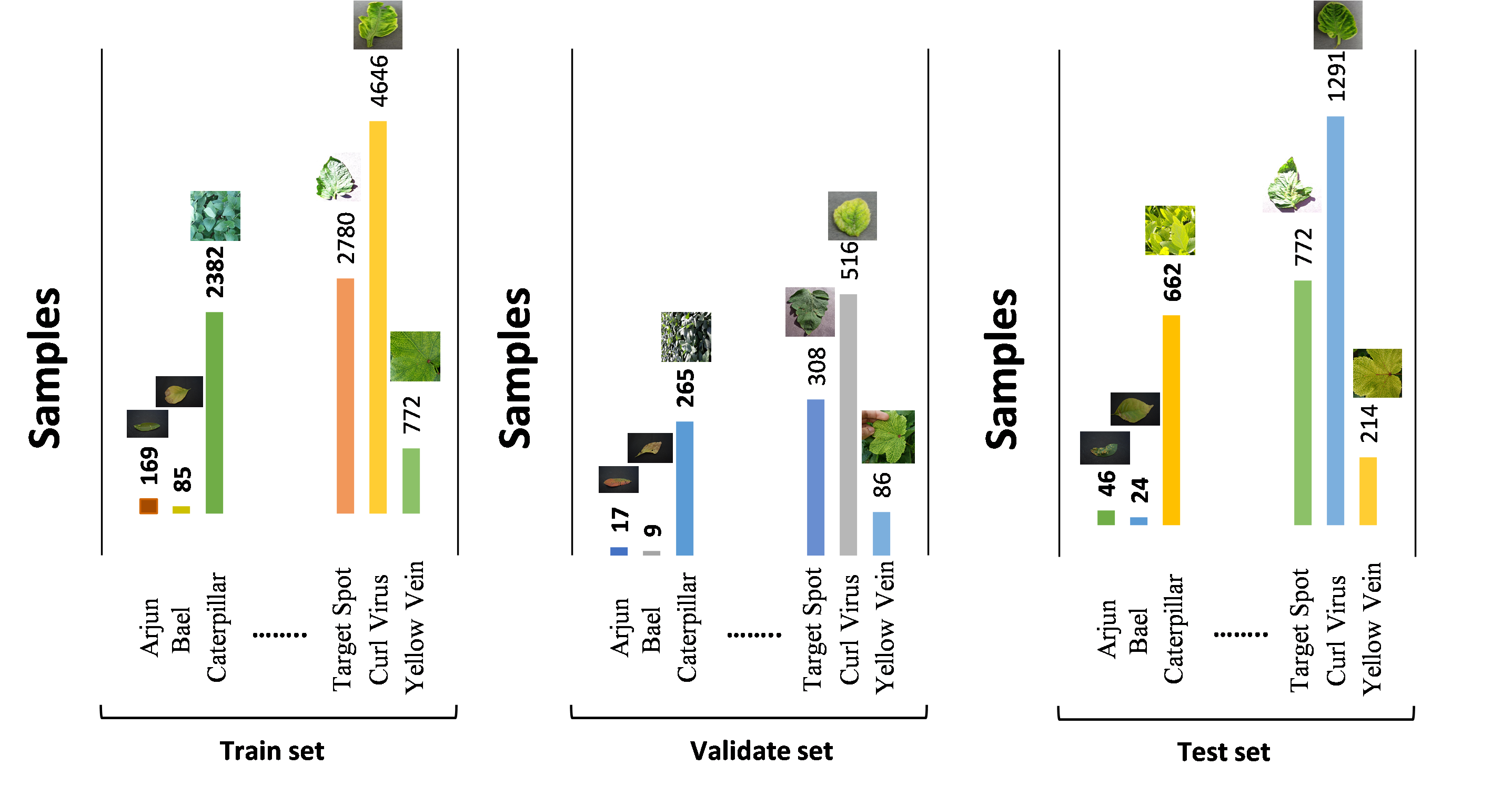}
    \caption{Ensuring Generalizability: A 70:10:20 Dataset Split for the Novel FloraSyntropy Archive}
    \label{fig4}
\end{figure}

   In this study, a structured approach to dataset partitioning was employed to ensure robust model evaluation and prevent overfitting. The available data was initially split into a dedicated hold-out test set comprising 20\% of the entire dataset. This test set was completely sequestered and used only for the final, unbiased evaluation of the model's performance on unseen data after all training and tuning was complete. The remaining 80\% of the data was designated as the training pool. From this pool, a validation set was then carved out, constituting 10\% of the original total dataset (which equates to 70\% of the training pool). This validation set is crucial for tuning hyperparameters, making architectural decisions like the Deep Block, and providing ongoing performance metrics during the training process without leaking information from the test set. This proposed ratio of 70:10:20 (Train: Validation: Test) is highly efficient as it allocates a substantial majority of the data for training the complex model, while still reserving a statistically significant portion for reliable final testing. The explicit separation of the validation set from the training data ensures that the model's generalization is monitored on data it hasn't been directly optimized on, leading to a more accurate and trustworthy assessment of its real-world capability. Fig \ref{fig4} illustrate the dataset distribution.
\subsection{Evaluating FloraSyntropy-Net framework procedure}
To evaluate the performance of the FloraSyntropy-Net framework, we carried out a thorough analysis using evaluation criteria frequently applied in evaluation assignments. The outcomes supported the validity of our strategy and were in line with findings from earlier research \cite{asim2020robust,nabeel2023dna}. Four categories TP (True Positive), TN (True Negative), FP (False Positive), and FN (False Negative) were assigned to each plant diagnosis sample. In order to determine performance metrics like accuracy (EQU-18), precision (EQU -19), recall (EQU -20), and F1-score (EQU-21), which offer a numerical understanding of the model effectiveness in differentiating between correct and incorrect diagnoses, the performance evaluation was carried out using these classifications and the corresponding mathematical formulations.
\begin{equation}
\text{Accuracy} = \frac{TP + TN}{TP + FP + TN + FN} \quad \text{(EQU-18)}
\end{equation}

\begin{equation}
\text{Precision} = \frac{TP}{TP + FP} \quad \text{(EQU-19)}
\end{equation}

\begin{equation}
\text{Recall} = \frac{TP}{TP + FN} \quad \text{(EQU-20)}
\end{equation}

\begin{equation}
F1 \, \text{Score} = \frac{2 \times (\text{Precision} \times \text{Recall})}{\text{Precision} + \text{Recall}} \quad \text{(EQU-21)}
\end{equation}
\section{Result and discussion}
This section details the methodology and experimental validation of the FloraSyntropy-Net model. We begin by describing the Novel FloraSyntropy Archive: Dataset Preparation and the Running Setup and Hyperparameters. Adherence to the TRIPOD Checklist ensures standardized reporting. The model's performance is then rigorously evaluated through a Large-Scale Plant Disease Classification benchmark and compared against existing CNN and SOTA models. Further analysis includes an Agricultural Application: Interpretable Analysis, a Domain Adaptation Test, and a critical Ablation study to deconstruct the model's architecture.
\subsection{Novel FloraSyntropy Archive: Dataset Preparation}
To guarantee robust performance, preprocessing of the globally gathered plant image is necessary prior to training the FloraSyntropy Framework. In order to improve the accuracy of diverse plant disease identification, this stage combines the various plant images from 13 publicly available agriculture dataset repositories. This enables the model to focus diverse plants on the regions that are irrelevant. All images are standardized to a uniform input size of 224 × 224 pixels in order to prevent bias and distortion brought on by differences in image dimensions. By ensuring uniform feature extraction across the 13 datasets, this standardization enhances FloraSyntropy generalization capabilities. Furthermore, pretreatment contributes to more consistent and dependable FloraSyntropy performance by reducing frequent problems with plant image, such as noise in image intensity.
   
\begin{figure}[h]
    \centering
    \includegraphics[width = 11cm]{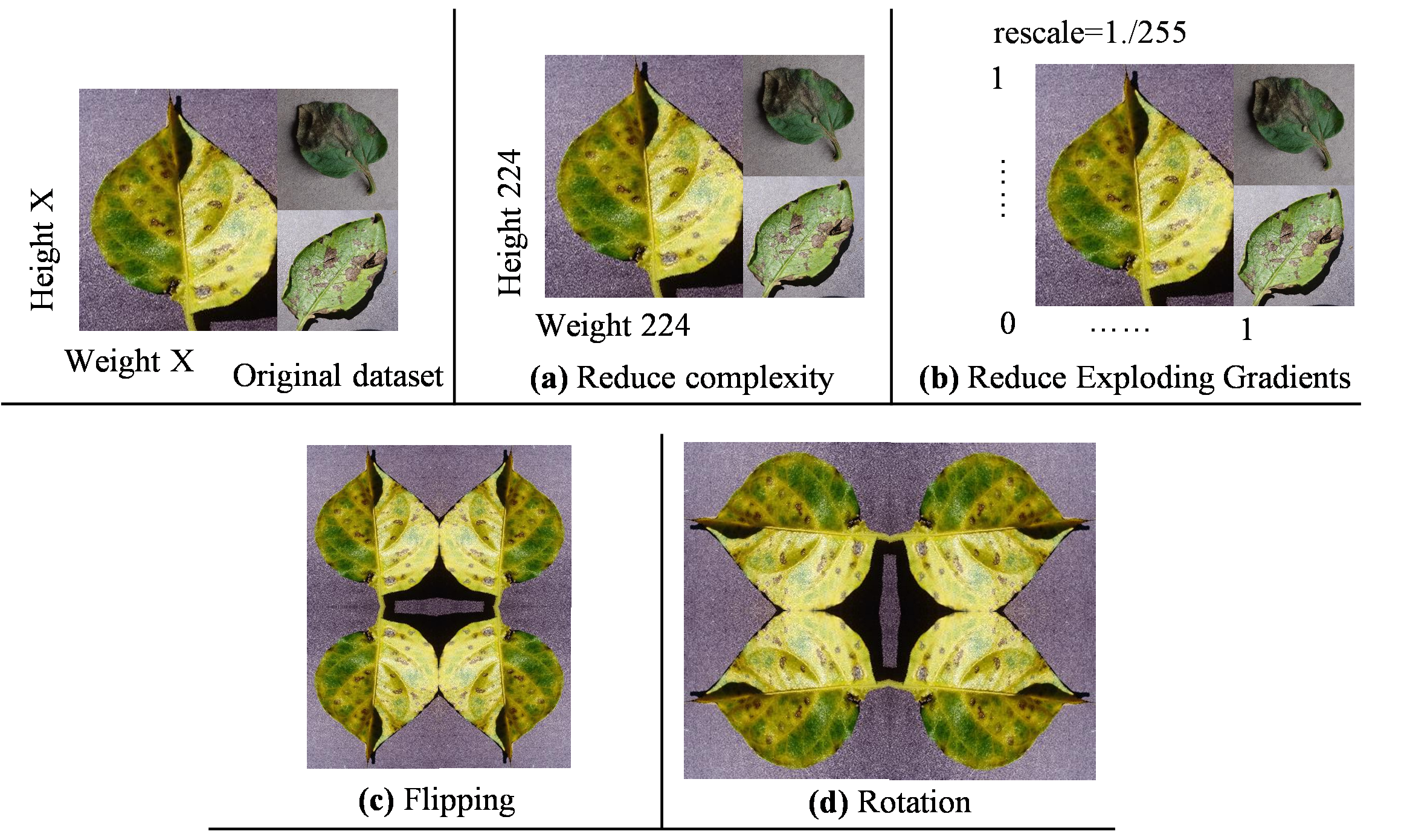}
    \caption{An overview of the preprocessed and augmented plant images: Novel FloraSyntropy Archive}
    \label{fig5}
\end{figure}

   To increase the diversity of our novel FloraSyntropy archive, we used three different augmentation techniques in this study. The distribution of plant image following dataset balancing, which guarantees fair representation across all categories, is shown in Table \ref{t1}. An illustration of the augmentation procedures is also provided by Fig \ref{fig5}, which shows the visual comparison between the original and augmented images. This method supports the model capacity to generalize successfully across many categories in addition to exposing it to a wider range of inputs. We reduced the possibility of skewed drawings that could result from class imbalance by maintaining equal class proportions during training, providing a more trustworthy indicator of the model capacity for generalization.
\begin{table}[!ht]
\centering
\caption{Distribution of the training and validation sets of large-scale plant images after dataset balancing}
\begin{tabular}{lrrlrr}
\hline
\textbf{Class} & \textbf{Train} & \textbf{Validation} & \textbf{Class} & \textbf{Train} & \textbf{Validation} \\
\hline
AlstoniaScholaris-Diseased & 4712 & 524 & Olive-AculusOlearius & 4712 & 524 \\
AlstoniaScholaris-Healthy & 4710 & 521 & Olive-Healthy & 4712 & 524 \\
Apple-BlackRot & 4712 & 524 & Olive-PeacockSpot & 4712 & 524 \\
Apple-CedarRust & 4712 & 524 & Orange-Haunglongbing & 4712 & 524 \\
Apple-Healthy & 4712 & 524 & Peach-BacterialSpot & 4712 & 524 \\
Apple-Scab & 4709 & 520 & Peach-Healthy & 4712 & 524 \\
Arjun-Diseased & 4712 & 524 & Pepper-BacterialSpot & 4707 & 519 \\
Arjun-Healthy & 4712 & 524 & Pepper-Healthy & 4712 & 524 \\
Bael-Diseased & 4712 & 524 & Pomegranate-Diseased & 4712 & 524 \\
Banana-Cordana & 4712 & 524 & Pomegranate-Healthy & 4712 & 524 \\
Banana-Healthy & 4712 & 524 & PongamiaPinnata-Diseased & 4712 & 524 \\
Banana-Pestalotiopsis & 4712 & 524 & PongamiaPinnata-Healthy & 4712 & 524 \\
Banana-Sigatoka & 4712 & 524 & Potato-EarlyBlight & 4712 & 524 \\
Basil-Healthy & 4712 & 524 & Potato-Healthy & 4712 & 524 \\
Betel-Diseased & 4712 & 524 & Potato-LateBlight & 4712 & 524 \\
Betel-Healthy & 4712 & 524 & Raspberry-Healthy & 4712 & 524 \\
Blueberry-healthy & 4712 & 524 & Rice-Blast & 4712 & 524 \\
Cassava-Bacterial Blight & 4712 & 524 & Rice-BrownSpot & 4712 & 524 \\
Cassava-Brown Streak Disease & 4712 & 524 & Rice-Healthy & 4712 & 524 \\
Cassava-Green Mottle & 4712 & 524 & Rice-Hispa & 4712 & 524 \\
Cassava-Healthy & 4712 & 524 & Soybean-Caterpillar & 4712 & 524 \\
Cassava-Mosaic Disease & 4712 & 524 & Soybean-DiabroticaSpeciosa & 4712 & 524 \\
Cherry-Healthy & 4712 & 524 & Soybean-Healthy & 4712 & 524 \\
Cherry-PowderyMildew & 4712 & 524 & Squash-PowderyMildew & 4712 & 524 \\
Chinar-Diseased & 4712 & 524 & Strawberry-Healthy & 4712 & 524 \\
Chinar-Healthy & 4712 & 524 & Strawberry-Scorch & 4712 & 524 \\
Coffe-Leaf & 4712 & 524 & Sugarcane-Diseased & 4712 & 524 \\
Corn-Cercospora & 4712 & 524 & Sugarcane-Healthy & 4712 & 524 \\
Corn-CommonRust & 4712 & 524 & Tea-Algal & 4712 & 524 \\
Corn-Healthy & 4712 & 524 & Tea-Anthracnose & 4712 & 524 \\
Corn-NorthernBlight & 4712 & 524 & Tea-BirdEye & 4712 & 524 \\
Cotton-BacterialBlight & 4712 & 524 & Tea-BrownBlight & 4712 & 524 \\
Cotton-CurlVirus & 4712 & 524 & Tea-GrayLight & 4712 & 524 \\
Cotton-FussariumWilt & 4712 & 524 & Tea-Healthy & 4712 & 524 \\
Cotton-Healthy & 4712 & 524 & Tea-RedSpot & 4712 & 524 \\
Gauva-Diseased & 4712 & 524 & Tea-WhiteSpot & 4712 & 524 \\
Gauva-Healthy & 4712 & 524 & Tomato-BacterialSpot & 4712 & 524 \\
Grape-BlackRot & 4712 & 524 & Tomato-EarlyBlight & 4712 & 524 \\
Grape-blight & 4712 & 524 & Tomato-Healthy & 4712 & 524 \\
Grape-Esca & 4712 & 524 & Tomato-LateBlight & 4712 & 524 \\
Grape-Healthy & 4712 & 524 & Tomato-Mold & 4712 & 524 \\
Jamun-Diseased & 4712 & 524 & Tomato-MosaicVirus & 4712 & 524 \\
Jamun-Healthy & 4712 & 524 & Tomato-SeptoriaSpot & 4712 & 524 \\
Jatropha-Diseased & 4712 & 524 & Tomato-SpiderMites & 4712 & 524 \\
Jatropha-Healthy & 4712 & 524 & Tomato-TargetSpot & 4712 & 524 \\
Lemon-Diseased & 4712 & 524 & Tomato-YellowCurlVirus & 4712 & 524 \\
Lemon-Healthy & 4712 & 524 & YellowVein-Diseased & 4712 & 524 \\
Mango-Diseased & 4712 & 524 & YellowVein-Healthy & 4712 & 524 \\
Mango-Healthy & 4712 & 524 & & & \\
\hline
\end{tabular}
\label{t1}
\end{table}
\subsection{Running Setup}
The FloraSyntropy-Net framework was developed and trained using the Python programming. The Global-Net model architecture was built using the adaptable PyCharm framework, and Python offered a productive environment for the implementation and performance of many tasks. All tests were carried out in the Python environment to optimize speed and computational efficiency \cite{wasim2018multi}, taking advantage of a GPU capabilities to speed up the training and testing procedures. In particular, an NVIDIA RTX5040 Tesla GPU and 32 GB of RAM were used, guaranteeing enough hardware resources for the demanding tasks at hand. This configuration was essential for effectively managing the extensive data processing needed for the assessment of the model performance.
\subsection{TRIPOD Checklist: Ensuring Standardized Reporting in Plant Disease Diagnosis}
Verifying our comprehension of disease pathways requires accurate and open reporting of research. When diagnosing agricultural diseases, it is essential to follow established protocols, such as the TRIPOD checklist. Table \ref{TRO-P1},\ref{TRO-P2} presents the TRIPOD checklist \cite{khan2026} encourages consistency, dependability, and reproducibility in research findings by offering an organized framework for reporting prediction models. In the detection of plant diseases, adherence to reporting guidelines is important. Our study can guarantee that results are reliable and applicable to agricultural practice by increasing the transparency of model construction, validation, and interpretation. This section examines how the TRIPOD checklist is essential for improving decision-making, guaranteeing high-quality reporting, and boosting the dependability of diagnostic models.
\begin{table}
\centering
\caption{TRIPOD Checklist for FloraSyntropy-Net Framework (Part 1)}
\begin{tabular}{p{3cm}p{4cm}p{4cm}l}
\hline
\textbf{TRIPOD Section} & \textbf{Description} & \textbf{Our Mapping} & \textbf{Checklist} \\ \hline
\textbf{Title} & Identify the study as developing and/or validating a generalize prediction model, the target population, and the outcome to be predicted. & FloraSyntropy-Net: Scalable Deep Learning for Novel Plant Disease Classification identifies the model and its purpose. & Pass \\ \hline
\textbf{Abstract} & Provide a summary of objectives, study design, results, and conclusions. & Paper abstract summarizes the framework objectives, methodology (federated learning, optimization, finetuning), and key results. & Pass \\ \hline
\textbf{Introduction} &  &  &  \\ \hline
3a. \textbf{Background} & Explain the plant disease context and rationale for developing or validating the prediction model. & The Introduction explains the context of large-scale plant disease diagnosis and the limitations of existing methods. & Pass \\ 
3b. \textbf{Objectives} & Specify the objectives, including whether the study describes the development or validation of the model, or both. & Section 1.1. Study contribution explicitly states the objectives and novel contributions of the FloraSyntropy-Net framework. & Pass \\ \hline
\textbf{Related Work} & Review of current models, approaches, and pertinent studies, along with an analysis of their shortcomings. & Highlight the distinctive contributions of the work by contrasting FloraSyntropy-Net with earlier federated learning models and their use in plant disease detection. & Pass \\ \hline
\textbf{Methods} &  &  &  \\ \hline
4. \textbf{Source of data} & Describe the study design or source of data. & Section 3.1. Novel Dataset FloraSyntropy Archive and 4.1. Dataset Preprocessing describe the source, nature, and handling of the data. & Pass \\ 
5a. \textbf{Participants} & Specify key elements of the study setting and the inclusion/exclusion criteria. & Section 3.1. implies the dataset contains specific plant disease imagery. The criteria would be detailed in the dataset description. & Fair* \\ 
6a. \textbf{Outcome} & Clearly define the outcome that is predicted. & The outcome is plant disease classification, defined in Section 3.1. and 4.4. Large-Scale Plant Disease Classification Experiment. & Pass \\ 
7a. \textbf{Predictors} & Clearly define all predictors used in developing the final model. & Predictors are the image data from the FloraSyntropy Archive. The feature extraction process is defined in Section 3.2./3.3. (Global-Net selection \& finetuning). & Pass \\ 
8. \textbf{Sample size} & Explain how the study size was arrived at. & The dataset size and the 70:10:20 split (Section 3.1./4.1) justify the sample size used for training and evaluation. & Pass \\ 
\hline
\end{tabular}
\label{TRO-P1}
\end{table}

\begin{table}
\centering
\caption{TRIPOD Checklist for FloraSyntropy-Net Framework (Part 2)}
\begin{tabular}{p{3cm}p{4cm}p{4cm}l}
\hline
\textbf{TRIPOD Section} & \textbf{Description} & \textbf{Our Mapping} & \textbf{Checklist} \\ \hline
10. \textbf{Model development} & Specify the statistical methods used for model development. & Sections 3.2. (Memetic Algorithm), 3.3. (Finetune Architecture), and 3.4./3.5. /3.6. (Federated Learning) detail the machine learning methods for model development. & Pass \\ 
16. \textbf{Model performance} & Report performance measures for the prediction model. & Sections 4.4., 4.5., 4.6., and 4.7.4. are dedicated to reporting results like accuracy, precision, recall, F1-score, and statistical analysis. & Pass \\ \hline
\textbf{Results} &  &  &  \\ \hline
17. \textbf{Participants} & Describe the flow of participants and any deviations from the study plan. & Section 4.1. describes the dataset distribution and preprocessing, effectively describing the flow of data samples. & Fair* \\ 
19. \textbf{Interpretation} & Give an overall interpretation of the results. & Sections 4. (Result and discussion) and 5. Discussion provide interpretation of the results in the context of existing work and the study objectives. & Pass \\ 
20. \textbf{Implications} & Discuss the potential implications of the results. & Section 6. Conclusion and future direction discuss the implications and future work, suggesting the model's value and next steps. & Pass \\ \hline
21. \textbf{Discussion} & Discuss any limitations of the study. & Section 5. Discussion with limitation is explicitly dedicated to this purpose. & Pass \\ \hline
22. \textbf{Data and Code Availability} & To ensure transparency and repeatability, make the data and code used in the investigation available. & For reproducibility, make sure that the code is publicly available. & Fair* \\ \hline
23. \textbf{Reporting Compliance} & Adherence to reporting regulations, such as TRIPOD. & Our study complies with TRIPOD requirements by using the checklist for transparent reporting. & Pass \\ \hline
\end{tabular}
\label{TRO-P2}
\end{table}
\subsection{Large-Scale Plant Disease Classification Evaluation of Proposed model}
The integration of the Novel Deep Block into the FloraSyntropy-Net architecture yielded a clear and significant improvement in overall performance (Table \ref{PC-P1},\ref{PC-P2} (WO Novel Deep Block), and \ref{PC-P3},\ref{PC-P4} (With Novel Deep Block)), as evidenced by the increase in overall accuracy from 94.74\% to 96.38\%. This 1.64\% gain demonstrates that the block efficiently enhanced the model feature extraction capabilities, leading to more precise classification across the extensive dataset. A closer look at the per-class metrics reveals that this improvement was not uniform but targeted. Numerous challenging classes, particularly those with previously poor performance, saw substantial gains in F1-scores. For instance, the Arjun-Diseased class improved from 0.8261 to 0.9070, Bael-Diseased from 0.8627 to 0.9200, and Potato-LateBlight from 0.8544 to 0.9879. This pattern suggests the Novel Deep Block specifically improved the model ability to discern complex and subtle features that were previously difficult to classify, effectively addressing specific weaknesses in the base model.

  However, the upgrade was not universally beneficial for every single class. A comparative analysis shows slight regressions in a minority of cases, such as Rice-Blast (F1 dropping from 0.9285 to 9169) and Rice-Hispa (from 0.9280 to 0.8770). This indicates that while the new block optimizes the network for the vast majority of features, it may cause minor over-specialization or a shift in decision boundaries that slightly disadvantages a few specific classes. Despite these isolated regressions, the FloraSyntropy-Net effect is positive by implementing a FL approach with five client clones of our finely-tuned FloraSyntropy-Net model, which significantly enhance both the robustness and generalization capability of our model. This decentralized approach allows each client to learn specialized features from its unique subset of the data, while periodic aggregation of model weights on a central server synthesizes these diverse learnings into a single, robust global model. This process not only mitigates the risk of overfitting to specific data characteristics that may have caused minor regressions in individual classes during centralized training but also leverages collective learning from varied data distributions. 
  
Performance comparison of FloraSyntropy-Net against baseline models.

\begin{table}[!h]
\centering
\caption{Per Class Performance Comparison - FloraSyntropy-Net (Our) WO Novel Deep Block (Part 1)}
\begin{tabular}{p{2cm}cccp{2cm}cccc}
\hline
\textbf{Class} & \textbf{Precision} & \textbf{Recall} & \textbf{F1-Score} & \textbf{Class} & \textbf{Precision} & \textbf{Recall} & \textbf{F1-Score} & Overall\\
\hline
AlstoniaScholaris-Diseased & 0.8679 & 0.9020 & 0.8846 & Olive-AculusOlearius & 0.8652 & 0.6854 & 0.7649 & 94.74\\
AlstoniaScholaris-Healthy & 0.8571 & 0.8333 & 0.8451 & Olive-Healthy & 0.5054 & 0.9895 & 0.6690 & \\
Apple-BlackRot & 1.0000 & 0.9914 & 0.9957 & Olive-PeacockSpot & 0.9742 & 0.5171 & 0.6756 &\\
Apple-CedarRust & 0.9974 & 0.9873 & 0.9924 & Orange-Haunglongbing & 0.9985 & 0.9992 & 0.9989 &\\
Apple-Healthy & 0.9896 & 0.9955 & 0.9926 & Peach-BacterialSpot & 1.0000 & 0.9950 & 0.9975 &\\
Apple-Scab & 0.9865 & 0.9981 & 0.9923 & Peach-Healthy & 0.9854 & 0.9951 & 0.9902 & \\
Arjun-Diseased & 0.8261 & 0.8261 & 0.8261 & Pepper-BacterialSpot & 1.0000 & 0.9751 & 0.9874 &\\
Arjun-Healthy & 0.6029 & 0.9318 & 0.7321 & Pepper-Healthy & 0.9854 & 0.9951 & 0.9902 &\\
Bael-Diseased & 0.8148 & 0.9167 & 0.8627 & Pomegranate-Diseased & 1.0000 & 0.9751 & 0.9874 &\\
Banana-Cordana & 0.9310 & 0.8438 & 0.8852 & Pomegranate-Healthy & 0.9854 & 0.9951 & 0.9902 &\\
Banana-Healthy & 0.7879 & 1.0000 & 0.8814 & PongamiaPinnata-Diseased & 1.0000 & 0.9751 & 0.9874 & \\
Banana-Pestalotiopsis & 0.6531 & 0.9143 & 0.7619 & PongamiaPinnata-Healthy & 0.9940 & 0.9812 & 0.9876 & \\
Banana-Sigatoka & 0.8482 & 1.0000 & 0.9179 & Potato-EarlyBlight & 0.8548 & 0.9815 & 0.9138 &\\
Basil-Healthy & 1.0000 & 1.0000 & 1.0000 & Potato-Healthy & 0.9138 & 0.9298 & 0.9217 &\\
Betel-Diseased & 0.9474 & 0.8571 & 0.9000 & Potato-LateBlight & 0.9167 & 0.8000 & 0.8544 &\\
Betel-Healthy & 0.7368 & 0.9333 & 0.8235 & Raspberry-Healthy & 0.9091 & 0.9375 & 0.9231 &\\
Blueberry-healthy & 0.9913 & 0.9800 & 0.9856 & Rice-Blast & 0.9903 & 0.9868 & 0.9285 &\\
Cassava-Bacterial Blight & 0.5781 & 0.3978 & 0.4713 & Rice-BrownSpot & 0.9839 & 1.0000 & 0.9919 &\\
Cassava-Brown Streak Disease & 0.6827 & 0.6339 & 0.6574 & Rice-Healthy & 0.9841 & 0.9794 & 0.9818 &\\
Cassava-Green Mottle & 0.7229 & 0.6383 & 0.6780 & Rice-Hispa & 0.9959 & 0.9836 & 0.9280 &\\
Cassava-Healthy & 0.5593 & 0.7333 & 0.6346 & Soybean-Caterpillar & 0.6875 & 0.3929 & 0.5000 &\\
Cassava-Mosaic Disease & 0.6875 & 0.7765 & 0.7293 & Soybean-DiabroticaSpeciosa & 0.7778 & 0.2800 & 0.4118 &\\
\hline
\end{tabular}
\label{PC-P1}
\end{table}

\begin{table}
\centering
\caption{Per Class Performance Comparison - FloraSyntropy-Net (Our) WO Novel Deep Block (Part 2)}
\begin{tabular}{p{2cm}cccp{2cm}ccc}
\hline
\textbf{Class} & \textbf{Precision} & \textbf{Recall} & \textbf{F1-Score} & \textbf{Class} & \textbf{Precision} & \textbf{Recall} & \textbf{F1-Score} \\
\hline
Cherry-Healthy & 0.9973 & 0.9739 & 0.9855 & Soybean-Healthy & 0.5726 & 0.9579 & 0.7168 \\
Cherry-PowderyMildew & 0.9972 & 1.0000 & 0.9986 & Squash-PowderyMildew & 0.2400 & 0.1463 & 0.1818 \\
Chinar-Diseased & 0.9500 & 0.7917 & 0.8636 & Strawberry-Healthy & 0.8149 & 0.4456 & 0.5762 \\
Chinar-Healthy & 1.0000 & 0.9524 & 0.9756 & Strawberry-Scorch & 0.6304 & 0.8549 & 0.7257 \\
Coffe-Leaf & 0.9872 & 0.9904 & 0.9888 & Sugarcane-Diseased & 0.8473 & 0.9920 & 0.9139 \\
Corn-Cercospora & 0.8654 & 1.0000 & 0.9278 & Sugarcane-Healthy & 1.0000 & 0.9987 & 0.9994 \\
Corn-CommonRust & 0.9975 & 0.9817 & 0.9895 & Tea-Algal & 0.9987 & 0.9987 & 0.9987 \\
Corn-Healthy & 0.9773 & 1.0000 & 0.9885 & Tea-Anthracnose & 0.9986 & 1.0000 & 0.9993 \\
Corn-NorthernBlight & 0.9880 & 0.8906 & 0.9368 & Tea-BirdEye & 0.9875 & 0.9080 & 0.9461 \\
Cotton-BacterialBlight & 0.9730 & 0.8000 & 0.8780 & Tea-BrownBlight & 0.9286 & 0.5652 & 0.7027 \\
Cotton-CurlVirus & 0.9861 & 0.8452 & 0.9103 & Tea-GrayLight & 0.7778 & 0.3500 & 0.4828 \\
Cotton-FussariumWilt & 0.7835 & 0.9048 & 0.8398 & Tea-Healthy & 0.5000 & 0.8000 & 0.6154 \\
Cotton-Healthy & 0.9615 & 0.2941 & 0.4505 & Tea-RedSpot & 0.8182 & 0.3913 & 0.5294 \\
Gauva-Diseased & 1.0000 & 0.4643 & 0.6341 & Tea-WhiteSpot & 0.6667 & 0.6000 & 0.6316 \\
Gauva-Healthy & 0.9487 & 0.6727 & 0.7872 & Tomato-BacterialSpot & 1.0000 & 1.0000 & 1.0000 \\
Grape-BlackRot & 0.9925 & 0.9863 & 0.9894 & Tomato-EarlyBlight & 0.8235 & 0.9655 & 0.8889 \\
Grape-blight & 0.9975 & 0.9926 & 0.9950 & Tomato-Healthy & 0.5400 & 0.9659 & 0.9794 \\
Grape-Esca & 0.9943 & 0.9943 & 0.9943 & Tomato-LateBlight & 0.9689 & 0.9177 & 0.9426 \\
Grape-Healthy & 1.0000 & 1.0000 & 1.0000 & Tomato-Mold & 0.9988 & 0.9659 & 0.9794 \\
Jamun-Diseased & 0.8571 & 0.7826 & 0.8182 & Tomato-MosaicVirus & 0.9689 & 0.9177 & 0.9426 \\
Jamun-Healthy & 0.8594 & 0.9821 & 0.9167 & Tomato-SeptoriaSpot & 0.9988 & 0.9952 & 0.9970 \\
Jatropha-Diseased & 0.9167 & 0.8800 & 0.8980 & Tomato-SpiderMites & 0.9410 & 0.9779 & 0.9591 \\
Jatropha-Healthy & 0.8929 & 0.9259 & 0.9091 & Tomato-TargetSpot & 0.9340 & 0.9925 & 0.9624 \\
Lemon-Diseased & 0.7000 & 0.4667 & 0.5600 & Tomato-YellowCurlVirus & 0.8215 & 0.9079 & 0.8625 \\
Lemon-Healthy & 0.7619 & 1.0000 & 0.8649 & YellowVein-Diseased & 0.1404 & 0.0281 & 0.0468 \\
Mango-Diseased & 0.9792 & 0.8868 & 0.9307 & YellowVein-Healthy & 0.9880 & 0.9973 & 0.9927 \\
Mango-Healthy & 0.8684 & 0.9706 & 0.9167 & & & & \\
\hline
\end{tabular}
\label{PC-P2}
\end{table}

\begin{table}
\centering
\caption{Per Class Performance Comparison - FloraSyntropy-Net (Our) With Novel Deep Block (Part 1)}
\begin{tabular}{p{2cm}cccp{2cm}ccccc}
\hline
\textbf{Class} & \textbf{Precision} & \textbf{Recall} & \textbf{F1-Score} & \textbf{Class} & \textbf{Precision} & \textbf{Recall} & \textbf{F1-Score} & Overall \\
\hline
AlstoniaScholaris-Diseased & 0.8727 & 0.9412 & 0.9057 & Olive-AculusOlearius & 0.8365 & 0.7472 & 0.7893  &96.38\\
AlstoniaScholaris-Healthy & 0.9032 & 0.7778 & 0.8358 & Olive-Healthy & 0.5207 & 0.9947 & 0.6835 \\
Apple-BlackRot & 1.0000 & 0.9951 & 0.9975 & Olive-PeacockSpot & 0.9735 & 0.5034 & 0.6637 \\
Apple-CedarRust & 0.9987 & 0.9861 & 0.9923 & Orange-Haunglongbing & 1.0000 & 0.9992 & 0.9996 \\
Apple-Healthy & 0.9925 & 0.9925 & 0.9925 & Peach-BacterialSpot & 1.0000 & 1.0000 & 1.0000 \\
Apple-Scab & 0.9942 & 0.9981 & 0.9961 & Peach-Healthy & 0.9919 & 0.9951 & 0.9935 \\
Arjun-Diseased & 0.9750 & 0.8478 & 0.9070 & Pepper-BacterialSpot & 1.0000 & 0.9876 & 0.9937 \\
Arjun-Healthy & 0.7451 & 0.8636 & 0.8000 & Pepper-Healthy & 0.9964 & 0.9753 & 0.9857 \\
Bael-Diseased & 0.8846 & 0.9583 & 0.9200 & Pomegranate-Diseased & 0.9464 & 0.9815 & 0.9636 \\
Banana-Cordana & 0.9143 & 1.0000 & 0.9552 & Pomegranate-Healthy & 0.9032 & 0.9825 & 0.9412 \\
Banana-Healthy & 0.9630 & 1.0000 & 0.9811 & PongamiaPinnata-Diseased & 0.9107 & 0.9273 & 0.9189 \\
Banana-Pestalotiopsis & 0.8095 & 0.9714 & 0.8831 & PongamiaPinnata-Healthy & 0.8986 & 0.9688 & 0.9323 \\
Banana-Sigatoka & 0.8407 & 1.0000 & 0.9135 & Potato-EarlyBlight & 0.9868 & 0.9904 & 0.9886 \\
Basil-Healthy & 0.9677 & 1.0000 & 0.9836 & Potato-Healthy & 0.9761 & 1.0000 & 0.9879 \\
Betel-Diseased & 0.8947 & 0.8095 & 0.8500 & Potato-LateBlight & 0.9867 & 0.9891 & 0.9879 \\
Betel-Healthy & 0.8667 & 0.8667 & 0.8667 & Raspberry-Healthy & 0.9973 & 1.0000 & 0.9986 \\
Blueberry-healthy & 0.9899 & 0.9843 & 0.9871 & Rice-Blast & 0.8970 & 0.8107 & 0.9169 \\
Cassava-Bacterial Blight & 0.6119 & 0.4409 & 0.5125 & Rice-BrownSpot & 0.9062 & 0.3867 & 0.5421 \\
Cassava-Brown Streak Disease & 0.7634 & 0.6339 & 0.6927 & Rice-Healthy & 0.6466 & 0.8551 & 0.7364 \\
Cassava-Green Mottle & 0.6429 & 0.6702 & 0.6562 & Rice-Hispa & 0.8303 & 0.8390 & 0.8770 \\
Cassava-Healthy & 0.5676 & 0.7778 & 0.6562 & Soybean-Caterpillar & 0.8318 & 0.5529 & 0.6642 \\
\hline
\end{tabular}
\label{PC-P3}
\end{table}

\begin{figure}[!h]
\centering
\begin{minipage}[]{0.46\textwidth}
  \centering
  \includegraphics[width = \textwidth]{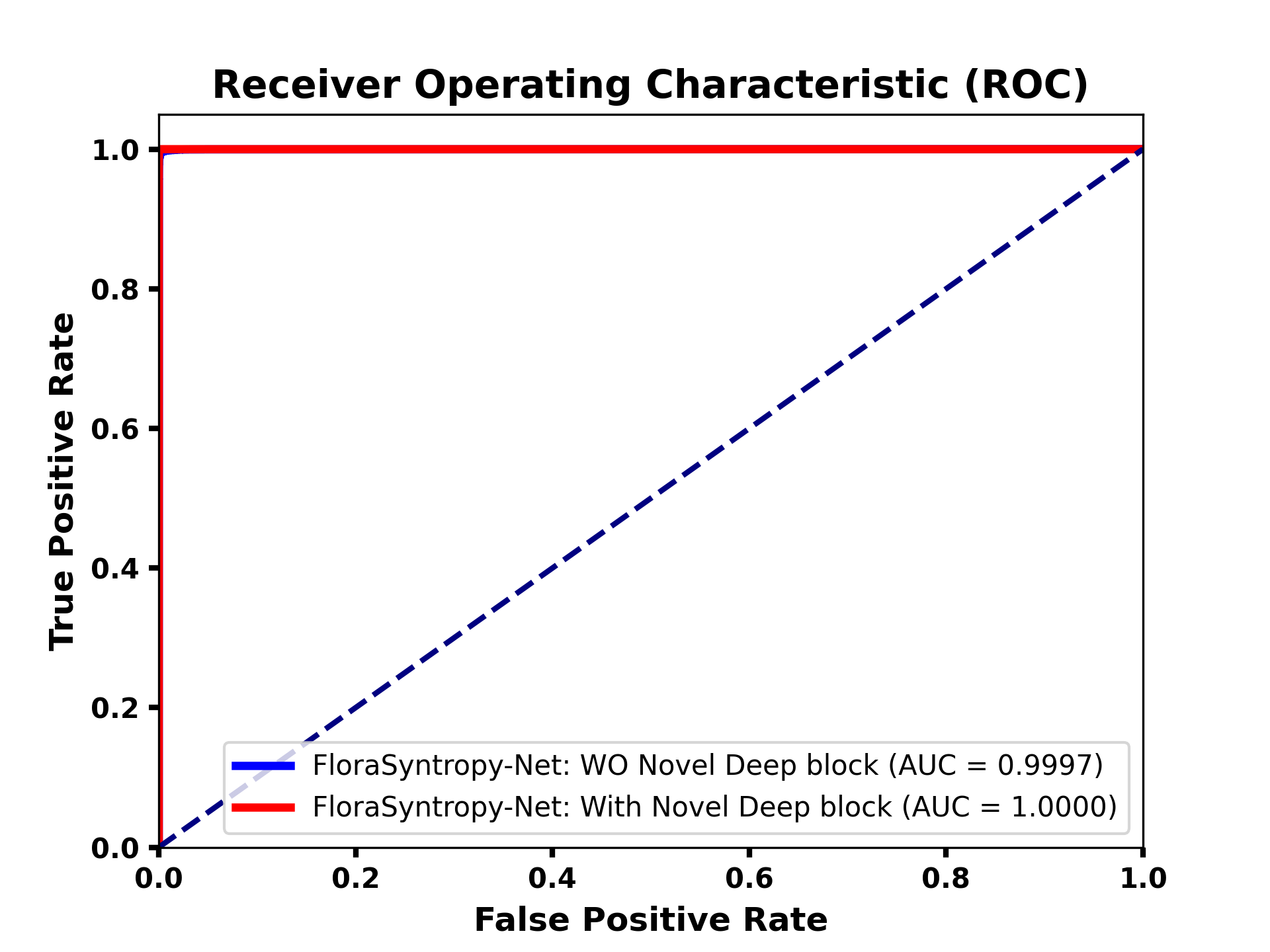}
  
    \begin{center}
     \textbf{a}    
     \end{center}
\end{minipage}
\begin{minipage}[]{0.46\textwidth}
  \centering
  \includegraphics[width = \textwidth]{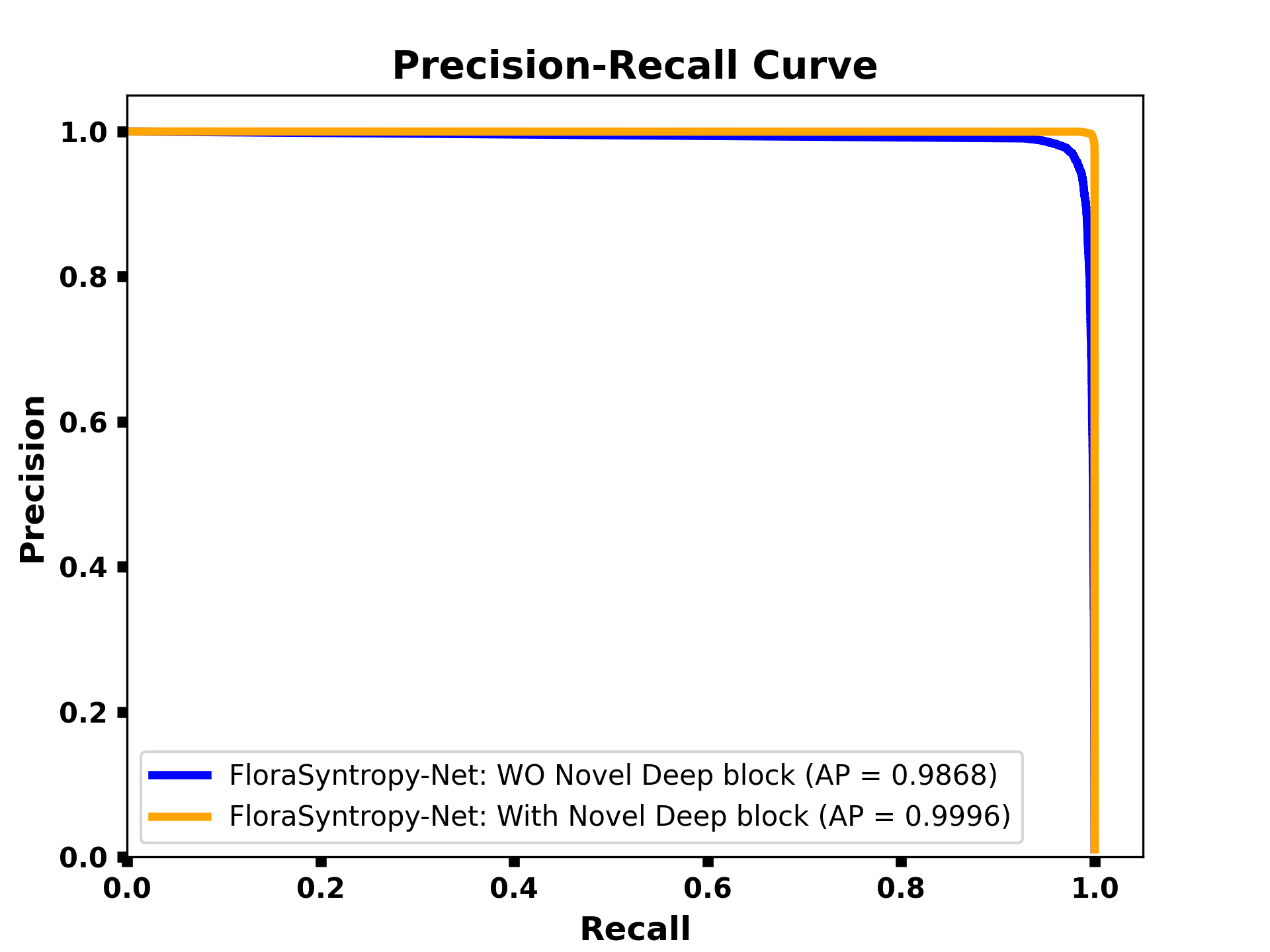}
 
     \begin{center}
     \textbf{b}    
     \end{center}
\end{minipage}
  \caption{A comparison of performance for the FloraSyntropy-Net model (a) ROC Curve (b) Precision-Recall Curve, demonstrating a significant reduction in misclassifications and an increase in diagonal accuracy with novel deep block.}
  \label{fig6}
\end{figure}

\begin{table}
\centering
\caption{Per Class Performance Comparison - FloraSyntropy-Net (Our) With Novel Deep Block (Part 2)}
\begin{tabular}{p{2cm}cccp{2cm}cccc}
\hline
\textbf{Class} & \textbf{Precision} & \textbf{Recall} & \textbf{F1-Score} & \textbf{Class} & \textbf{Precision} & \textbf{Recall} & \textbf{F1-Score} \\
\hline
Cassava-Mosaic Disease & 0.8056 & 0.6824 & 0.7389 & Soybean-DiabroticaSpeciosa & 0.6376 & 0.8617 & 0.7329 \\
Cherry-Healthy & 0.9973 & 0.9739 & 0.9855 & Soybean-Healthy & 0.8837 & 0.9940 & 0.9356 \\
Cherry-PowderyMildew & 1.0000 & 1.0000 & 1.0000 & Squash-PowderyMildew & 1.0000 & 0.9987 & 0.9994 \\
Chinar-Diseased & 1.0000 & 0.8333 & 0.9091 & Strawberry-Healthy & 0.9987 & 0.9987 & 0.9987 \\
Chinar-Healthy & 0.9524 & 0.9524 & 0.9524 & Strawberry-Scorch & 0.9919 & 1.0000 & 0.9960 \\
Coffe-Leaf & 0.9841 & 0.9904 & 0.9872 & Sugarcane-Diseased & 1.0000 & 0.9195 & 0.9581 \\
Corn-Cercospora & 0.9236 & 0.9840 & 0.9529 & Sugarcane-Healthy & 0.9884 & 1.0000 & 0.9942 \\
Corn-CommonRust & 0.9975 & 0.9792 & 0.9883 & Tea-Algal & 1.0000 & 0.4783 & 0.6471 \\
Corn-Healthy & 0.9974 & 1.0000 & 0.9987 & Tea-Anthracnose & 0.7500 & 0.4500 & 0.5625 \\
Corn-NorthernBlight & 0.9681 & 0.9495 & 0.9587 & Tea-BirdEye & 0.4571 & 0.8000 & 0.5818 \\
Cotton-BacterialBlight & 0.9762 & 0.9111 & 0.9425 & Tea-BrownBlight & 0.9000 & 0.7826 & 0.8372 \\
Cotton-CurlVirus & 0.9865 & 0.8690 & 0.9241 & Tea-GrayLight & 0.7692 & 0.5000 & 0.6061 \\
Cotton-FussariumWilt & 0.8280 & 0.9167 & 0.8701 & Tea-Healthy & 1.0000 & 1.0000 & 1.0000 \\
Cotton-Healthy & 0.9600 & 0.2824 & 0.4364 & Tea-RedSpot & 0.9032 & 0.9655 & 0.9333 \\
Gauva-Diseased & 0.8696 & 0.7143 & 0.7843 & Tea-WhiteSpot & 0.6222 & 1.0000 & 0.7671 \\
Gauva-Healthy & 0.9773 & 0.7818 & 0.8687 & Tomato-BacterialSpot & 0.9973 & 0.9711 & 0.9840 \\
Grape-BlackRot & 0.9962 & 0.9850 & 0.9906 & Tomato-EarlyBlight & 0.9487 & 0.9533 & 0.9510 \\
Grape-blight & 0.9988 & 0.9963 & 0.9975 & Tomato-Healthy & 0.9952 & 0.9928 & 0.9940 \\
Grape-Esca & 0.9887 & 0.9929 & 0.9908 & Tomato-LateBlight & 0.9660 & 0.9767 & 0.9713 \\
Grape-Healthy & 1.0000 & 1.0000 & 1.0000 & Tomato-Mold & 0.9555 & 0.9937 & 0.9742 \\
Jamun-Diseased & 0.9474 & 0.7826 & 0.8571 & Tomato-MosaicVirus & 0.8188 & 0.8884 & 0.8522 \\
Jamun-Healthy & 0.8615 & 1.0000 & 0.9256 & Tomato-SeptoriaSpot & 0.5796 & 0.5316 & 0.5452 \\
Jatropha-Diseased & 0.9167 & 0.8800 & 0.8980 & Tomato-SpiderMites & 0.9920 & 1.0000 & 0.9960 \\
Jatropha-Healthy & 0.8929 & 0.9259 & 0.9091 & Tomato-TargetSpot & 0.9831 & 0.9806 & 0.9818 \\
Lemon-Diseased & 1.0000 & 0.4000 & 0.5714 & Tomato-YellowCurlVirus & 0.9908 & 0.9961 & 0.9934 \\
Lemon-Healthy & 0.7442 & 1.0000 & 0.8533 & YellowVein-Diseased & 0.9861 & 0.9953 & 0.9907 \\
Mango-Diseased & 0.9444 & 0.9623 & 0.9533 & YellowVein-Healthy & 0.9887 & 1.0000 & 0.9943 \\
Mango-Healthy & 0.9189 & 1.0000 & 0.9577 &  & & & \\
\hline
\end{tabular}
\label{PC-P4}
\end{table}
Based on the comparative ROC and Precision-Recall curve analysis (Fig \ref{fig6}) conducted on the large-scale novel FloraSyntropy Archive, the FloraSyntropy-Net framework incorporating the novel deep block demonstrates a measurable performance enhancement, achieving a perfect AUC of 1.000 compared to 0.999 without the block, and an efficient Average Precision of 1.000 versus 0.999. This measurable improvement signifies a superior ability to maintain high true positive rates while minimizing false positives across 97 disease classes. The 0.001 gain in both metrics, though numerically small, reflects the elimination of the final marginal errors and confirms the model strengthened capacity for precise feature discrimination, leading to a marked reduction in inter-class confusion. The results validate that the architectural innovation introduced by the deep block is critically effective in enhancing classification robustness and reliability on complex, real-world plant disease imagery.
\subsection{Performance comparison of FloraSyntropy-Net with existing CNN models}
Our proposed FloraSyntropy-Net demonstrates a significant and substantial superiority over a suite of SOTA DL models across all key metrics. Established architectures like DenseNet201 and ResNet50V2 achieved competitive (Table \ref{t2}) but lower baseline accuracy (93.73\% and 93.48\% respectively), with their precision, recall, and F1-scores consistently remaining in the 0.86-0.89 range. This indicates a significant number of misclassifications that balance out in overall accuracy. In unambiguous contrast, FloraSyntropy-Net achieves a significantly higher overall accuracy of 96.38\% and delivers a superior balance between precision (0.9525) and recall (0.9409), culminating in an efficient F1-score of 0.9504. This performance gap, particularly in the F1-score which is critical for imbalanced datasets, underscores the efficacy of our novel architectural enhancements and training methodology in achieving more precise and reliable plant disease classification.
\begin{table}[ht]
\centering
\caption{Performance comparison of FloraSyntropy-Net against baseline models}
\begin{tabular}{lcccc}
\hline
\textbf{Model} & \textbf{Precision} & \textbf{Recall} & \textbf{F1-Score} & \textbf{Accuracy} \\
\hline
DenseNet121 & 0.8886 & 0.8722 & 0.8702 & 0.9306 \\
DenseNet169 & 0.8958 & 0.8791 & 0.8768 & 0.9318 \\
DenseNet201 & 0.8986 & 0.8864 & 0.8838 & 0.9373 \\
InceptionV3 & 0.8632 & 0.8433 & 0.8361 & 0.9334 \\
MobileNetV1 & 0.8890 & 0.8739 & 0.8704 & 0.9306 \\
MobileNetV2 & 0.8881 & 0.8731 & 0.8714 & 0.9283 \\
ResNet50V2 & 0.8904 & 0.8688 & 0.8700 & 0.9348 \\
ResNet101V2 & 0.8857 & 0.8673 & 0.8681 & 0.9248 \\
ResNet152V2 & 0.8843 & 0.8618 & 0.8638 & 0.9255 \\
VGG16 & 0.8124 & 0.7917 & 0.7880 & 0.8692 \\
VGG19 & 0.7931 & 0.7793 & 0.7751 & 0.8528 \\
FloraSyntropy-Net (Our) & 0.9525 & 0.9409 & 0.9504 & 0.9638 \\
\hline
\end{tabular}
\label{t2}
\end{table}
\subsection{Performance comparison of FloraSyntropy-Net with existing SOTA models}
The comparative analysis of plant disease classification methods, as summarized in the table \ref{t3}, demonstrates a clear progression in performance and methodological robustness, culminating in the proposed FloraSyntropy-Net framework. While earlier approaches such as SVM-based classifiers (Shrivastava et al.\cite{shrivastava2021}, 94.65\%; Sharif et al \cite{sharif2018}, 95.80\%) and hybrid autoencoders (Singh et al., 99.35\%) achieved high accuracy on smaller, domain-specific datasets like Plant Village or BananaLSD, their evaluations were limited to controlled environments, absent large-scale validation. In contrast, the FloraSyntropy-Net framework, leveraging the novel, globally representative FloraSyntropy Archive, achieves a competitive accuracy of 96.38\% while uniquely incorporating rigorous large-scale validation. This ensures not only high performance but also proven generalizability across diverse, real-world agricultural scenarios, addressing a critical gap in prior studies.
\begin{table}[ht]
\centering
\caption{Comparative analysis of plant disease classification methodologies, highlighting dataset scope, reported accuracy, and the critical presence of large-scale validation}
\begin{tabular}{lp{3cm}p{3cm}ll}
\hline
\textbf{Reference}                  & \textbf{Method}             & \textbf{Dataset Used}      & \textbf{Accuracy} & \textbf{Large-Scale} \\ \hline
Harakannanavar et al. \cite{harakannanavar2022}          & ML-Classifier               & Plant Village (Tomato Leaf) & 88.00\%           & Absent                           \\ 
Shrivastava et al. \cite{shrivastava2021}             & SVM-Classifier              & Indira Gandhi               & 94.65\%           & Absent                           \\ 
Shah et al. \cite{shah2022}                    & CSVM                        & Plant Village               & More than 90.00\% & Absent                           \\ 
Singh et al. \cite{singh2020}                   & HAE                         & BananaLSD                   & 99.35\%           & Absent                           \\ 
Sharif et al. \cite{sharif2018}                  & M-SVM                       & Cotton Leaf                 & 95.80\%           & Absent                           \\ 
Proposed (Our)                      & FloraSyntropy-Net framework & FloraSyntropy-Net Archive   & 96.38\%           & Presents                         \\ \hline
\end{tabular}
\label{t3}
\end{table}
\subsection{Agricultural Application: Interpretable Analysis}
To ensure the FloraSyntropy-Net framework is not merely a high-performing black box but a reliable and trustworthy tool for agricultural diagnostics, this section provides an in-depth interpretable analysis. We employ a suite of explainable AI techniques to visually and statistically decipher the model decision-making process. Specifically, GRADCAM visualizations are used to highlight the discriminative image regions that most influenced the model predictions. Together, these methods provide critical insights into the model behavior, validate its focus on biologically relevant features, and ultimately build confidence in its practical deployment for real-world agricultural applications.
\subsection*{GRADCAM Visualization}
\begin{figure}[h]
    \centering
    \includegraphics[width = 10cm]{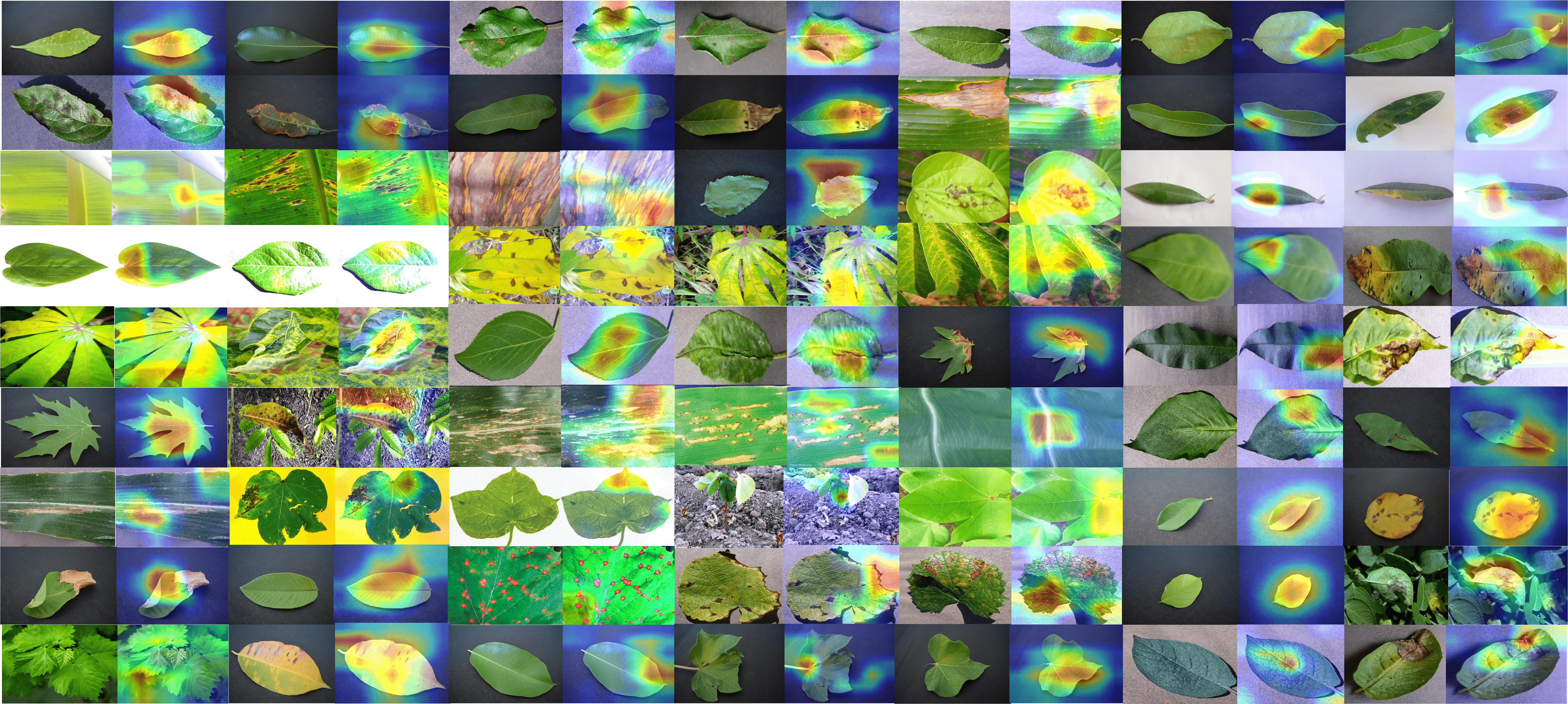}
    \caption{Overview of GRADCAM visualizations for FloraSyntropy-Net to highlight the discriminative image regions that most influenced the predictions demonstrating a significant reduction in misclassifications and an increase in diagonal interpretability with novel deep block.}
    \label{fig7}
\end{figure}

The GRADCAM visualizations \cite{khan2025,khan2026} for FloraSyntropy-Net consistently localize precise disease-specific features, such as lesions and insect damage, within plant imagery. This demonstrates the model ability to focus on diagnostically relevant regions rather than irrelevant background noise. The high-resolution heatmaps align perfectly with expert-agreed areas of infection, validating the model learned representations. This precise visual explainability confirms FloraSyntropy-Net efficacy as an accurate and trustworthy diagnostic tool. Fig \ref{fig7} presents the GRADCAM visualizations for FloraSyntropy-Net framework.
\subsection{Domain Adaptation Test: From Plants to Pests Classification}
Conducting a domain adaptation test, where a FloraSyntropy-Net model trained for plant disease classification is evaluated on the distinct task of pest classification, is a critical and rigorous method for validating true model robustness and generalization. This test moves beyond standard evaluation on similar data, intentionally introducing a significant domain shift in features from textural patterns of diseased leaves to the morphological features of insects. Its importance lies in stress-testing the model feature extraction capabilities; a model that performs well under such disparate conditions demonstrates that it has learned fundamental, transferable visual representations rather than merely memorizing dataset-specific artifacts. 
    
\begin{figure}[h]
    \centering
    \includegraphics[width = 12cm]{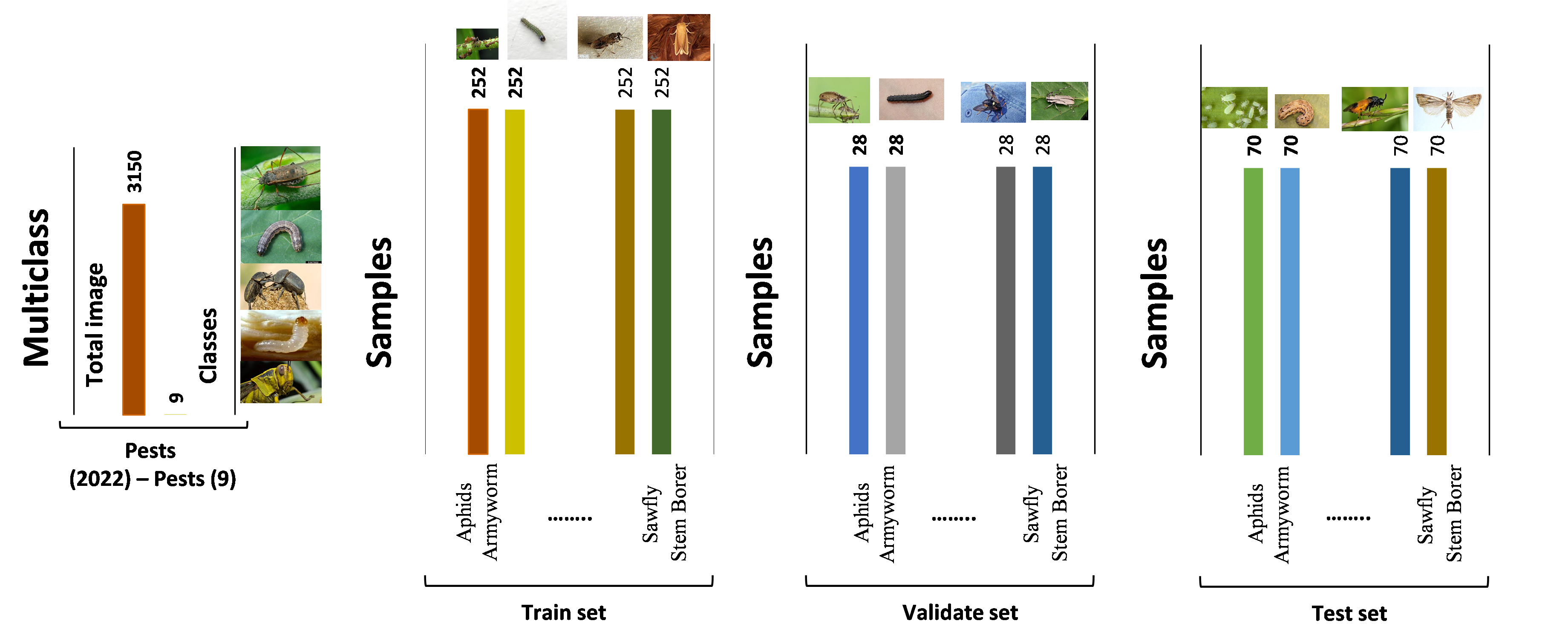}
    \caption{Representative dataset sample images and class distribution: Additional dataset}
    \label{fig8}
\end{figure}

\begin{figure}[!h]
\centering
\begin{minipage}[]{0.46\textwidth}
  \centering
  \includegraphics[width = \textwidth]{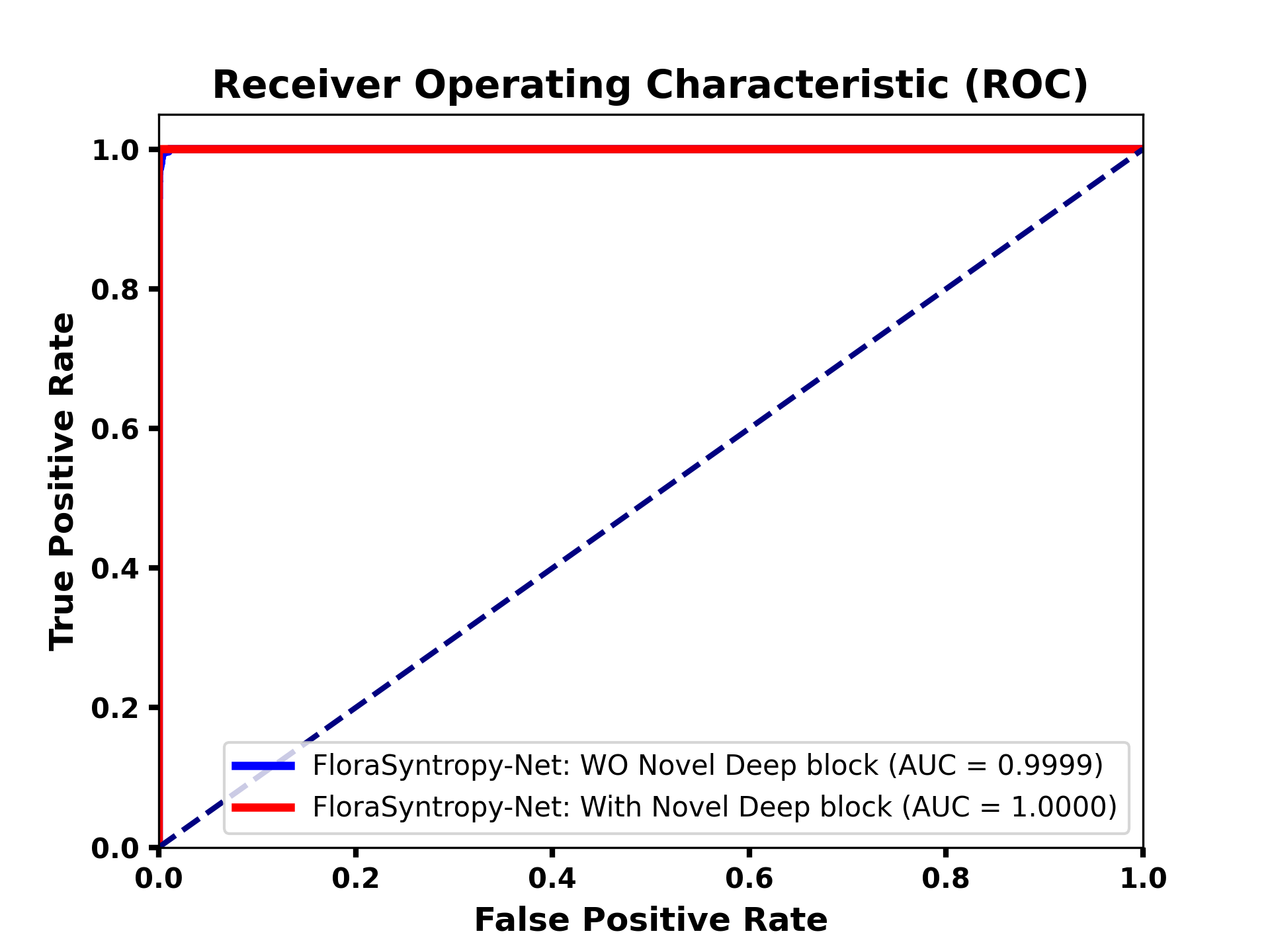}
  
    \begin{center}
     \textbf{a}    
     \end{center}
\end{minipage}
\begin{minipage}[]{0.46\textwidth}
  \centering
  \includegraphics[width = \textwidth]{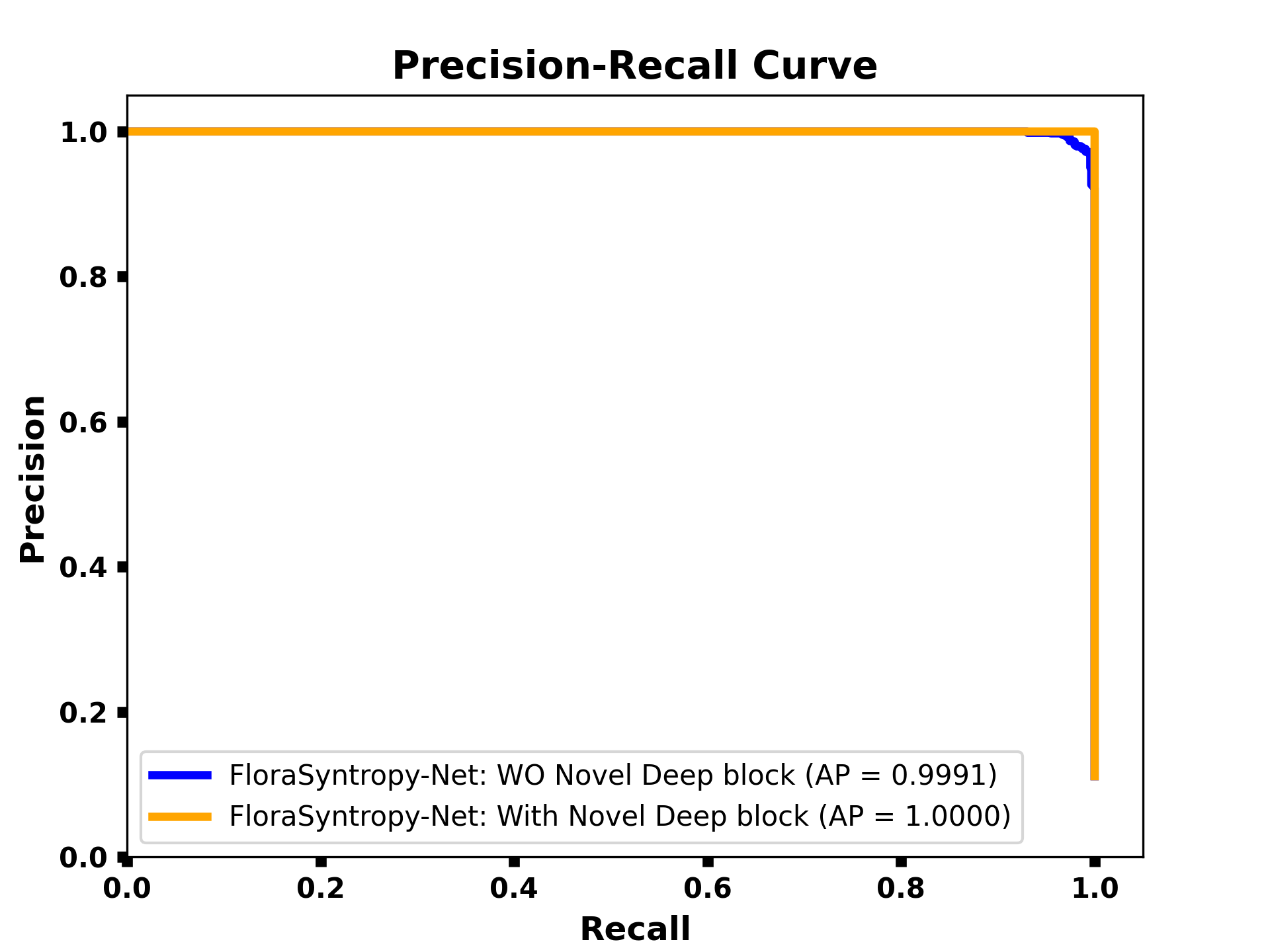}
 
     \begin{center}
     \textbf{b}    
     \end{center}
\end{minipage}
  \caption{A comparison of performance for the FloraSyntropy-Net model (a) ROC Curve (b) Precision-Recall Curve, demonstrating a significant reduction in misclassifications and an increase in diagonal accuracy with novel deep block: Additional dataset.}
  \label{fig9}
\end{figure}

\begin{figure}[!h]
\centering
\begin{minipage}[]{0.46\textwidth}
  \centering
  \includegraphics[width = \textwidth]{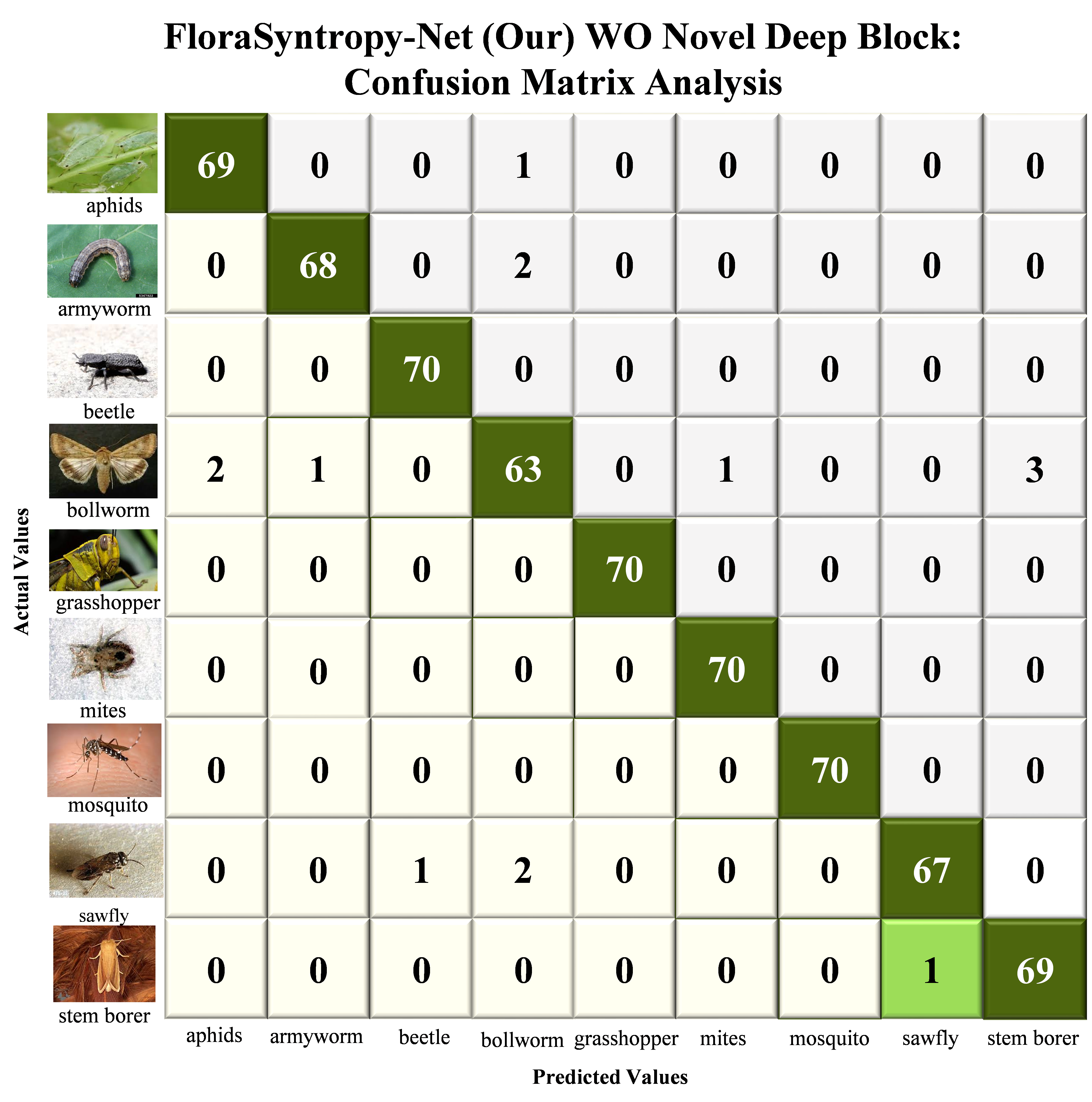}
  
    \begin{center}
     \textbf{a}    
     \end{center}
\end{minipage}
\begin{minipage}[]{0.46\textwidth}
  \centering
  \includegraphics[width = \textwidth]{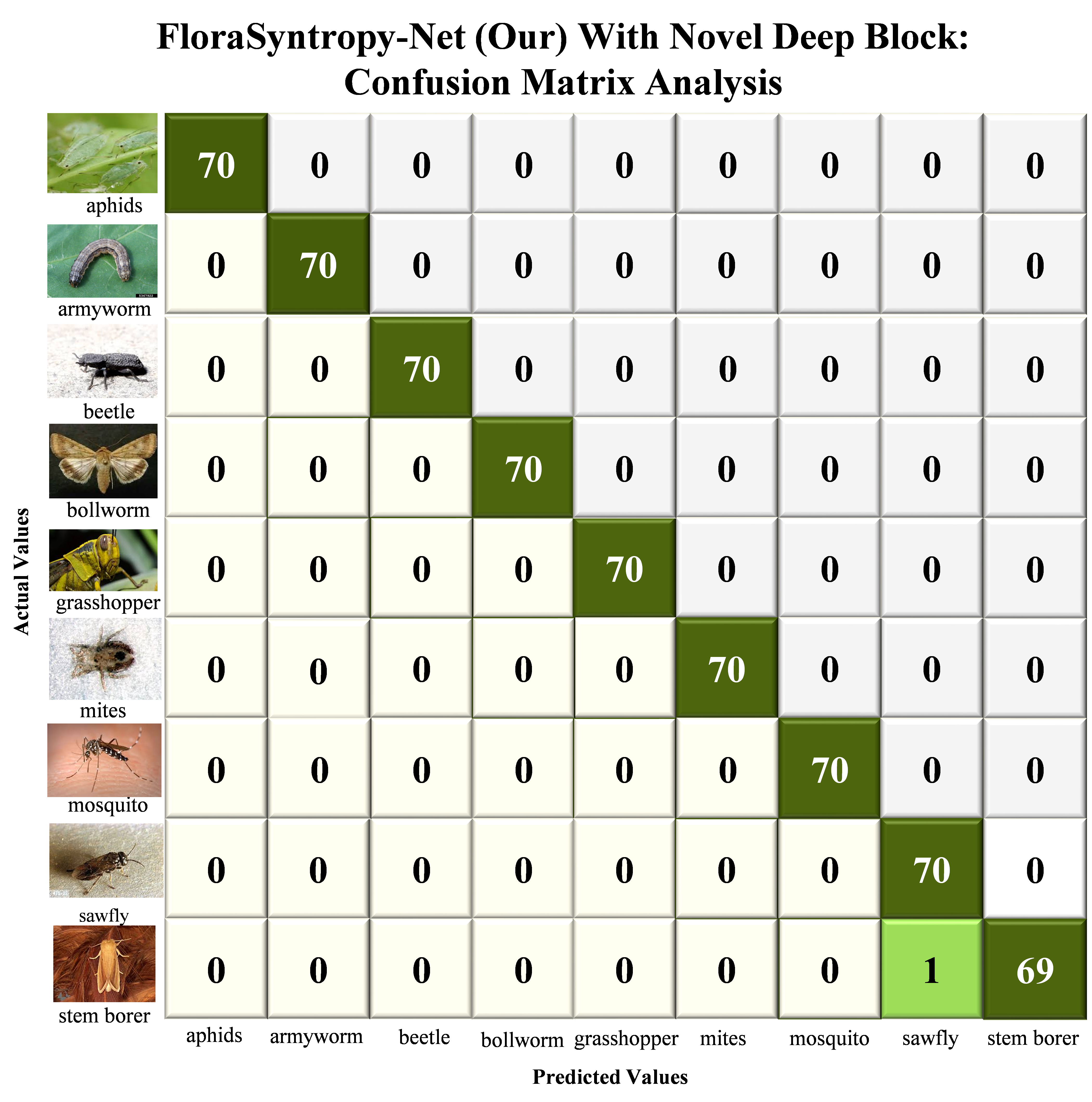}
 
     \begin{center}
     \textbf{b}    
     \end{center}
\end{minipage}
  \caption{A comparison of confusion matrices for the FloraSyntropy-Net model (a) without and (b) with the novel deep block, demonstrating a significant reduction in misclassifications and an increase in diagonal accuracy.}
  \label{fig10}
\end{figure}

\begin{figure}[h]
    \centering
    \includegraphics[width = 10cm]{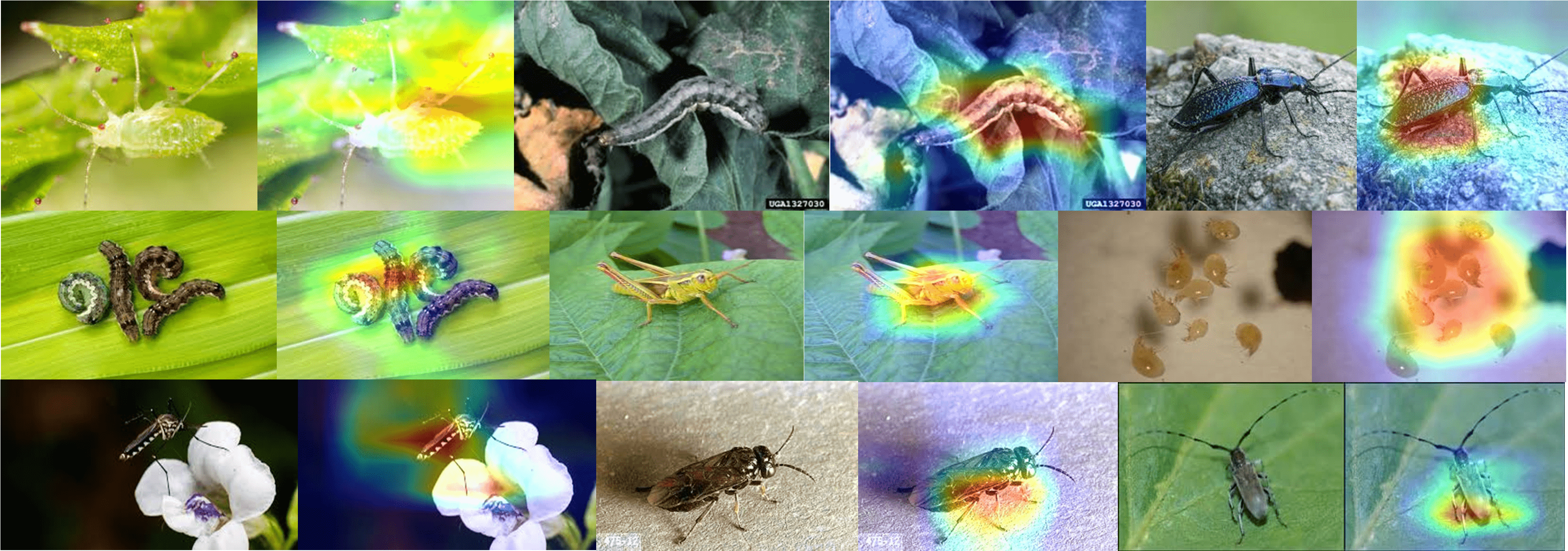}
    \caption{Overview of GRADCAM visualizations for FloraSyntropy-Net to highlight the discriminative image regions that most influenced the predictions demonstrating a significant reduction in misclassifications and an increase in diagonal interpretability with novel deep block: Additional dataset.}
    \label{fig11}
\end{figure}

\begin{figure}[!h]
\centering
\begin{minipage}[]{0.46\textwidth}
  \centering
  \includegraphics[width = \textwidth]{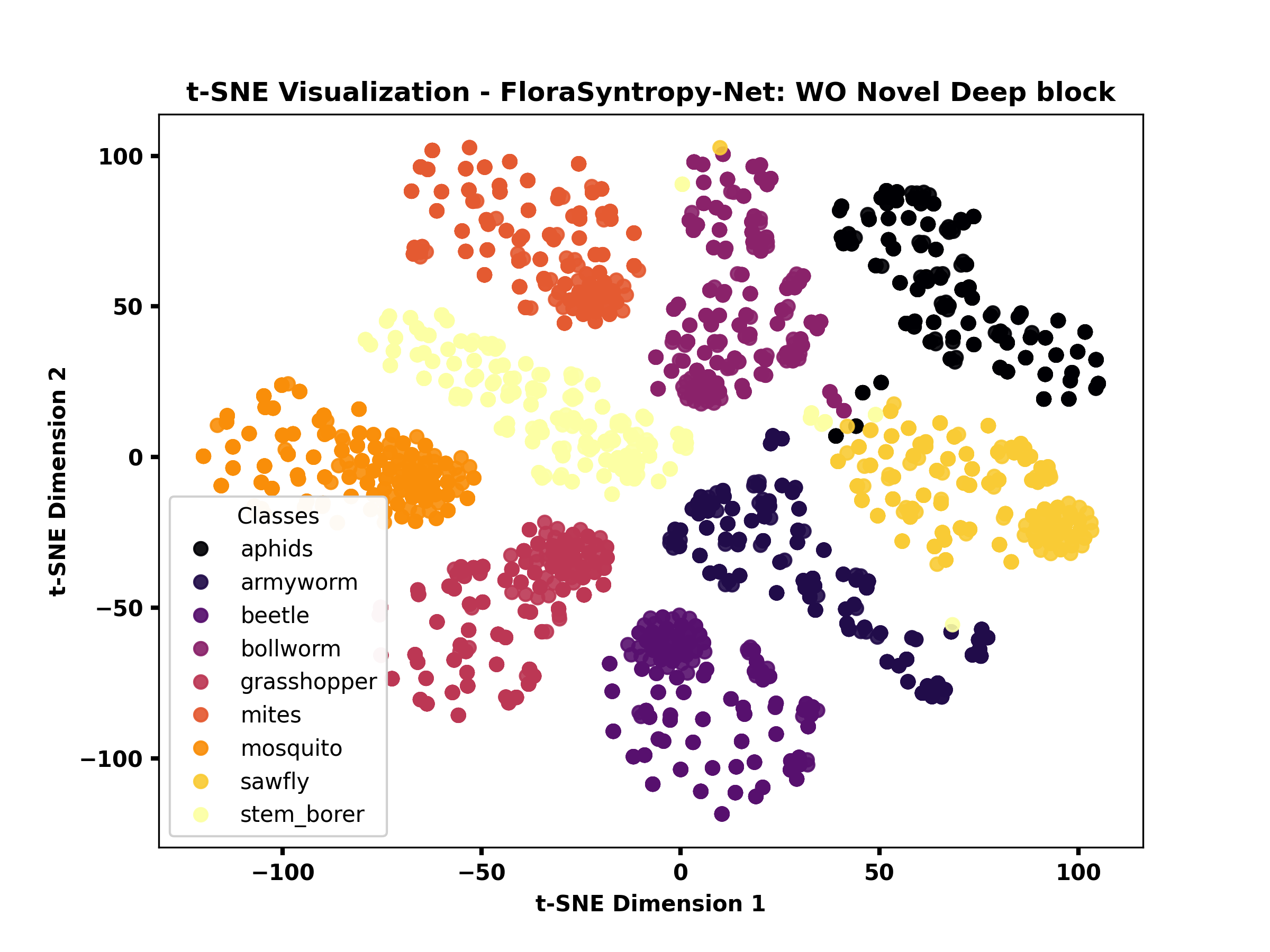}
  
    \begin{center}
     \textbf{a}    
     \end{center}
\end{minipage}
\begin{minipage}[]{0.46\textwidth}
  \centering
  \includegraphics[width = \textwidth]{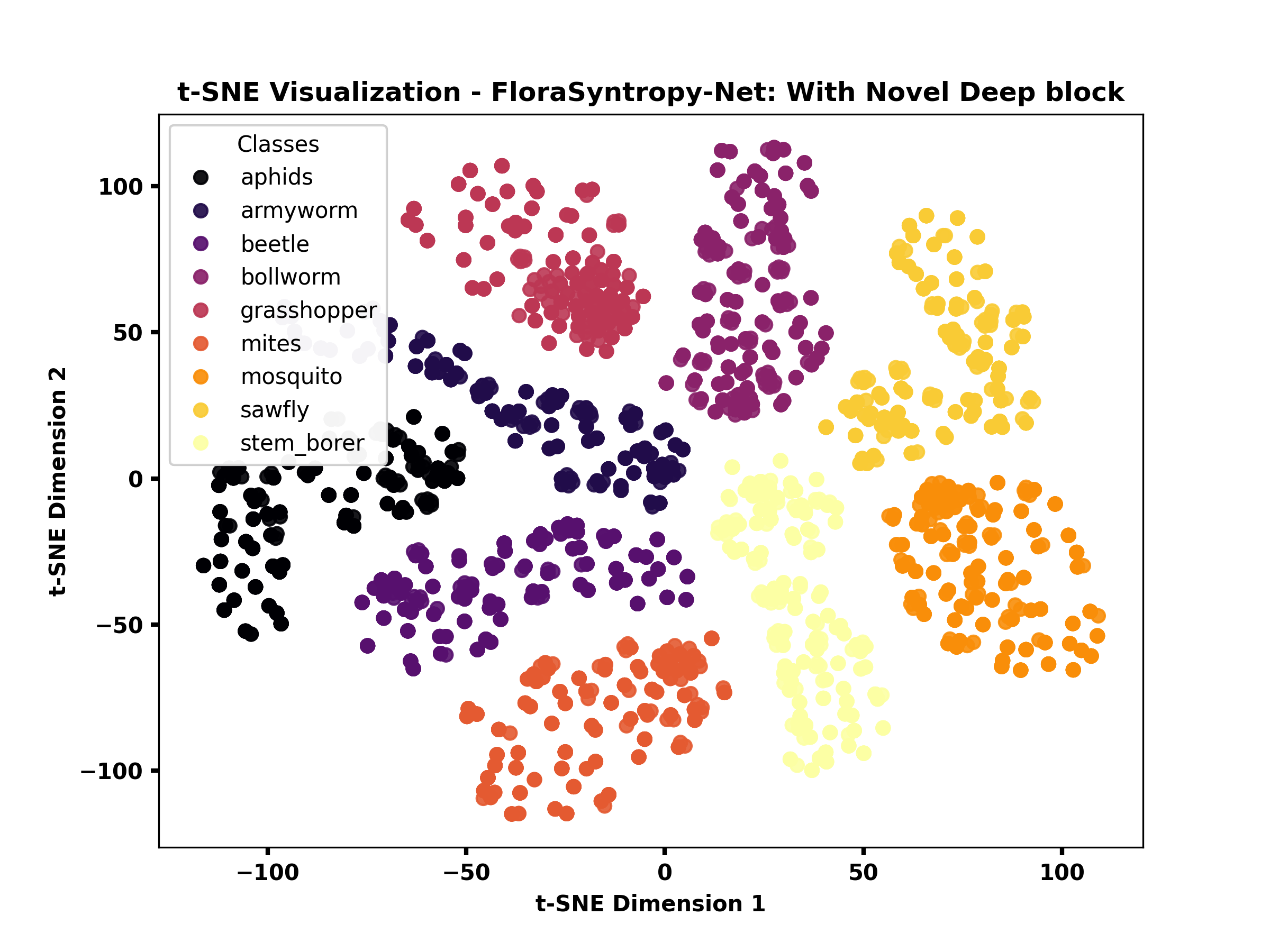}
 
     \begin{center}
     \textbf{b}    
     \end{center}
\end{minipage}
  \caption{A comparison of distinct clusters for each disease class in the latent space learned by FloraSyntropy-Net model (a) without and (b) with the novel deep block, demonstrating a significant reduction in misclassifications and an increase in diagonal accuracy.}
  \label{fig12}
\end{figure}

    In this study, we have used the multiclass Pest \cite{rao2025} dataset, which contains a diverse range of insect categories that pose distinct agricultural threats. This dataset provides an ideal benchmark to evaluate the cross-domain generalization capability of our plant-based model when validate with entirely different visual features and biological contexts. Fig \ref{fig8} showcases a representative sample of images from the dataset, illustrating the visual diversity and characteristics of the classes, alongside a detailed distribution chart that highlights the composition and balance of the data across categories.
The additional dataset test presents that integration of the novel deep block into the FloraSyntropy-Net architecture resulted in a substantial performance improvement, as evidenced by the comparative class-wise metrics (Table \ref{t4}). The most significant impact was observed on the Bollworm class, which saw its F1-score leap from 0.9130 to a perfect 1.0000, indicating the block enhanced ability to capture complex features that were previously challenging. This enhancement also driven the overall accuracy from 0.9778 to an exceptional 0.9984. Furthermore, while the without (deep block) model already performed well on several classes due to collective knowledge, the model (with novel deep block) achieved perfection (1.0000 across precision, recall, and F1) for the majority of classes, including Aphids, Armyworm, and Beetle, demonstrating that the deep block not only improved weaknesses but also boosted the model consistency and confidence across the entire dataset.
\begin{table}[h!]
\centering
\caption{Per class performance comparison of FloraSyntropy-Net Framework O/W Novel Deep block: Pest Dataset}
\begin{tabular}{ccccc}
\hline
\textbf{Class} & \textbf{Precision} & \textbf{Recall} & \textbf{F1-Score} & \textbf{Overall Accuracy} \\
\hline
\multicolumn{5}{c}{\textbf{FloraSyntropy-Net (Our) WO Novel Deep Block}} \\
\hline
Aphids          & 0.9718            & 0.9857          & 0.9787            & 0.9778                \\
Armyworm        & 0.9855            & 0.9714          & 0.9784            &                       \\
Beetle          & 0.9859            & 1.0000          & 0.9929            &                       \\
Bollworm        & 0.9265            & 0.9000          & 0.9130            &                       \\
Grasshopper     & 1.0000            & 1.0000          & 1.0000            &                       \\
Mites           & 0.9859            & 1.0000          & 0.9929            &                       \\
Mosquito        & 1.0000            & 1.0000          & 1.0000            &                       \\
Sawfly          & 0.9853            & 0.9571          & 0.9710            &                       \\
Stem\_borer     & 0.9583            & 0.9857          & 0.9718            &                       \\
\hline
\multicolumn{5}{c}{\textbf{FloraSyntropy-Net (Our) with Novel Deep Block}} \\
\hline
Aphids          & 1.0000            & 1.0000          & 1.0000            & 0.9984                \\
Armyworm        & 1.0000            & 1.0000          & 1.0000            &                       \\
Beetle          & 1.0000            & 1.0000          & 1.0000            &                       \\
Bollworm        & 1.0000            & 1.0000          & 1.0000            &                       \\
Grasshopper     & 1.0000            & 1.0000          & 1.0000            &                       \\
Mites           & 1.0000            & 1.0000          & 1.0000            &                       \\
Mosquito        & 1.0000            & 1.0000          & 1.0000            &                       \\
Sawfly          & 0.9859            & 1.0000          & 0.9929            &                       \\
Stem\_borer     & 1.0000            & 0.9857          & 0.9928            &                       \\
\hline
\end{tabular}
\label{t4}
\end{table}
Based on the comparative analysis of the ROC and Precision-Recall curves in Fig \ref{fig9}, the proposed FloraSyntropy-Net model with the novel deep block demonstrates near-perfect discriminative ability and classification reliability, outperforming the model without the deep block. The ROC curve shows an improvement in AUC from an already exceptional 0.9999 to a perfect 1.0000, indicating a perfect ability to distinguish between all classes without error. Similarly, the Precision-Recall curve prove this superior performance, with the Average Precision (AP) score increasing from 0.9991 to 1.0000, signifying that the model with the deep block achieves robust precision across all recall levels. This errorless performance on both metrics underscores that the novel deep block eliminates the remaining marginal errors, resulting in a model with maximum possible diagonal accuracy and zero misclassifications on the additional dataset.
A comparative analysis of the two confusion matrices (O/W novel deep block) clearly demonstrates the performance enhancement achieved by integrating the novel deep block into the FloraSyntropy-Net architecture (Fig \ref{fig10}). The baseline model (a), lacking this block, exhibits significant misclassifications, particularly confusing armyworm with other categories such as 2 incorrect class with bollworm and erroneously predicting mice as mosquito or stem borer. In contrast, the enhanced model (b) shows a substantial reduction in off-diagonal entries, with predictions becoming heavily concentrated on the correct diagonal classes. The only incorrect error is a single instance of mosquito being misclassified as mice. This reduction in confusion, especially for previously problematic classes like armyworm, beetle, and grasshopper, underscores the deep block efficacy in refining the model feature extraction and discriminatory capabilities. Consequently, the inclusion of the novel deep block yields a marked improvement in overall classification accuracy and model reliability.
Based on the GRADCAM visualizations in Fig \ref{fig11}, the proposed FloraSyntropy-Net model equipped with the novel deep block demonstrates an efficient and more precise interpretability analysis. The visualizations shows that the inclusion of the deep block enables the model to more accurately focus its attention on the most discriminative and pathologically relevant regions of the input images, such as specific leaf lesions or insect damage sites. This heightened spatial precision in activation maps directly correlates with the model significant reduction in misclassifications, as it is leveraging more meaningful visual features for its predictions. Consequently, the novel deep block not only boosts quantitative performance but also substantially increases the model transparency and diagonal interpretability, providing more trustworthy and actionable insights for plant disease diagnosis.
Figure 12, t-SNE projection \cite{khan2026} reveals well-separated, distinct clusters for each disease class in the latent space learned by FloraSyntropy-Net. This clear separation indicates the model efficient capability to extract and discriminate between highly nuanced pathological features. The minimal overlap between clusters provides statistical evidence of the model's high classification accuracy and generalization power. Thus, the visualization quantitatively proves the superiority of the proposed architecture's feature representation learning. In fig \ref{fig12}, the proposed FloraSyntropy-Net model with the novel deep block (b) demonstrates an improvement in comparative performance over the model without it (a). The incorporation of the deep block results in significantly more distinct and well-separated clusters for each disease class in the latent feature space, indicating that the model learns more discriminative and class-specific representations. This enhanced separation directly correlates with a reduction in misclassifications and an increase in diagonal accuracy, as the model can more effectively distinguish between visually similar conditions such as alpha arrhythmia, benign syndrome, and grasshopper infection. The clear clustering observed in (b) confirms that the novel architectural enhancement successfully addresses feature ambiguity, leading to superior classification robustness and accuracy.
\subsection{Ablation study}
In this section, we present a comprehensive ablation study to rigorously evaluate the individual contributions of our selected Global-Net backbone architecture and the novel Deep Block integrated within the FloraSyntropy-Net framework. The study systematically dissects the model by incrementally adding or removing key components comparing the performance of the baseline Global-Net (DenseNet201) alone, the Global-Net enhanced with our proposed Deep Block, and alternative configurations using established architectural blocks such as residual (ResNet), Inception, Naïve, and other connections in place of our design. This analysis not only quantifies the performance gain attributable solely to the novel Deep Block through metrics such as accuracy, F1-score, and precision but also validates its robustness and superiority over existing structural alternatives. By isolating each element, we demonstrate that the performance improvements are indeed a result of our FL collective knowledge architectural innovations, thereby reinforcing the design choices and providing empirical evidence of the efficacy and necessity of the proposed Deep Block within the overall system.
   
\begin{figure}[h]
    \centering
    \includegraphics[width = 10cm]{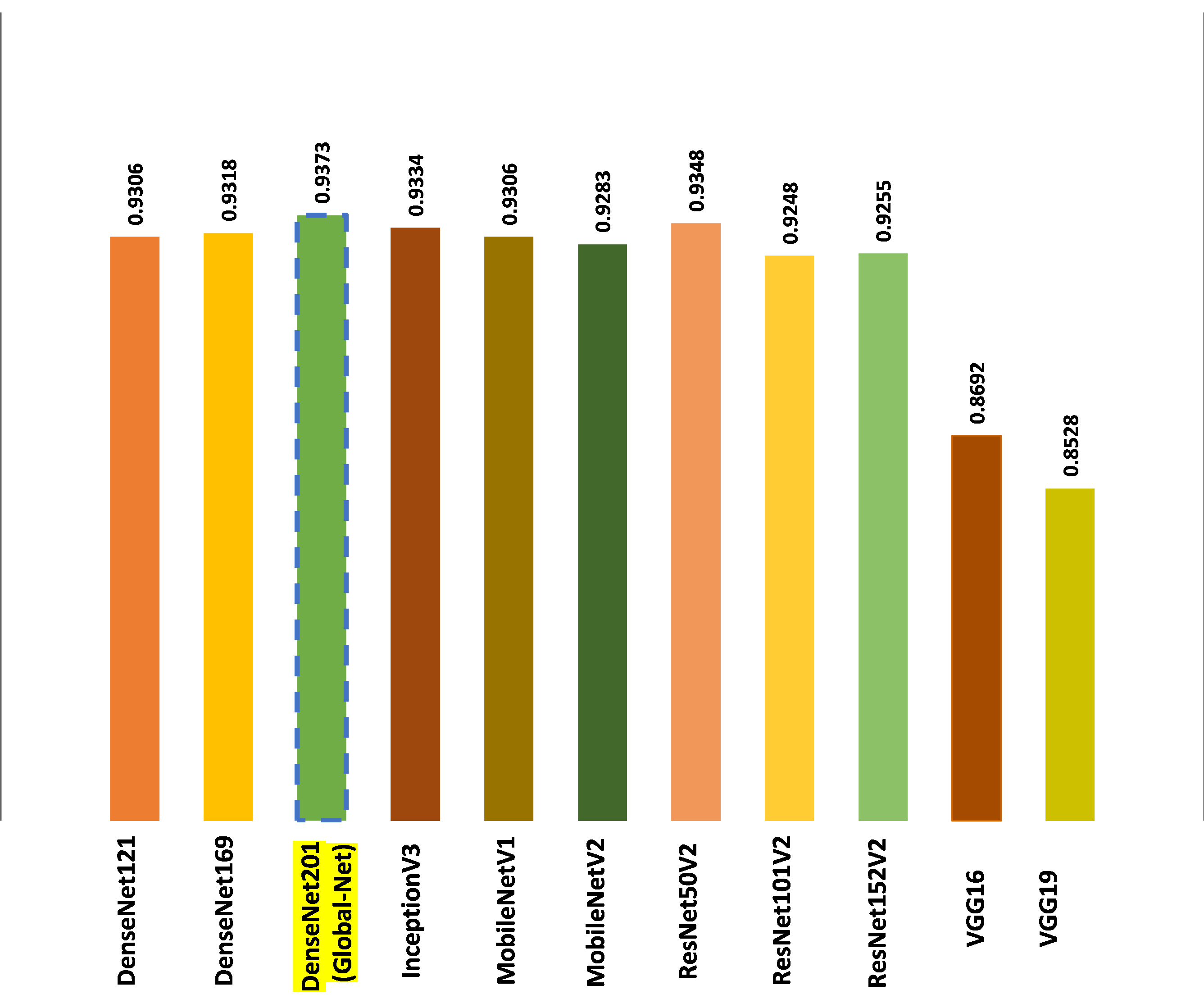}
    \caption{Comparative performance of various DL architectures. DenseNet201 (denoted as Global-Net and highlighted with a blue dashed border) achieved the highest performance (0.9373), outperforming all other evaluated models.}
    \label{fig13}
\end{figure}

   Fig \ref{fig13} illustrates a comparative analysis of model performance, evaluating architectures such as DenseNet variants and other convolutional networks based on accuracy. Among all models, DenseNet201 (referred to as Global-Net in this study) achieves superior performance with a value of 0.9373, as definitely denoted by its distinct blue dashed border. The remaining models, while still demonstrating strong and competitive results within the narrow band of 0.8692 to 0.9306, consistently fall short of the benchmark set by Global-Net. This performance gap underscores the efficacy of DenseNet201 as the optimal backbone feature extractor, approved its selection as the Global-Net within the proposed framework.
Table \ref{t5} presents the ablation study results demonstrate the clear superiority of the proposed Novel Deep Block over other prominent architectural blocks when integrated into the FloraSyntropy-Net framework. While all alternative blocks including Residual, Inception, Squeeze-and-Excitation (SE), Convolutional Block Attention Module (CBAM), and Coordinate Attention (CA) delivered highly competitive and comparable performance with accuracy scores range between 94.20\% and 94.48\%, the model equipped with the Novel Deep Block consistently outperformed them across every single metric. It achieved the highest scores in Precision (0.8925), Recall (0.8709), F1-Score (0.8704), Accuracy (94.74\%), and Cohen's Kappa (0.9463). This uniform dominance across all evaluation criteria presents that the Novel Deep Block provides a more effective mechanism for feature refinement and representation learning, leading to superior discriminatory power and the most robust overall performance.
\begin{table}[h!]
\centering
\caption{Ablation study comparing the performance of various architectural blocks integrated into the FloraSyntropy-Net framework, demonstrating the superior performance of the proposed Novel Deep Block across all metrics.}
\begin{tabular}{p{3cm}ccccc}
\hline
\textbf{Blocks} & \textbf{Precision} & \textbf{Recall} & \textbf{F1-Score} & \textbf{Accuracy} & \textbf{Cohen's K} \\ \hline
FloraSyntropy-Net with Residual Block & 0.8758 & 0.8593 & 0.8559 & 94.31\% & 0.9419 \\ 
FloraSyntropy-Net with Inception Block & 0.8755 & 0.8541 & 0.8519 & 94.27\% & 0.9414 \\ 
FloraSyntropy-Net with SE & 0.8864 & 0.8657 & 0.8658 & 94.20\% & 0.9407 \\ \hline
FloraSyntropy-Net with CBAM & 0.8835 & 0.8581 & 0.8599 & 94.39\% & 0.9427 \\ 
FloraSyntropy-Net with Naïve Inception & 0.8841 & 0.8601 & 0.8605 & 94.48\% & 0.9436 \\ 
FloraSyntropy-Net with CA & 0.8844 & 0.8574 & 0.8593 & 94.40\% & 0.9428 \\ \
FloraSyntropy-Net with Novel Deep Block & 0.8925 & 0.8709 & 0.8704 & 94.74\% & 0.9463 \\ \hline
\end{tabular}
\label{t5}
\end{table}
\section{Discussion with limitation}
The results of this study demonstrate that the proposed FloraSyntropy-Net framework represents a significant advancement in the development of robust and generalizable AI models for plant disease diagnosis. The integration of the novel Deep Block was contributory in enhancing feature extraction, leading to a measurable performance gain of 1.64\% in overall accuracy and resolving specific weaknesses in classifying challenging disease classes. Moreover, the FloraSyntropy-Net framework federated design, supported by a weighted aggregation strategy, successfully leveraged heterogeneous data from multiple clients to build a global model that is inherently more robust than one trained on isolated, small-scale datasets. The most compelling evidence of this robustness is the model exceptional performance (99.84\% accuracy) on the cross-domain Pest 2022 dataset. This demonstrates an unprecedented ability to generalize beyond its training domain, effectively learning universal visual features that are applicable to entirely new classification tasks involving different biological subjects. This suggests that FloraSyntropy-Net successfully mitigates the generalization gap that has plagued previous studies \cite{acidri2020,deng2022} reliant on smaller, less diverse data.

   Despite its strong performance, this study is not without limitations: 
\begin{itemize}
    \item The FL process, while advantageous for privacy and data diversity, is computationally intensive and requires significant coordination between participating clients (e.g., agriculture research institutes). The synchronization of model updates across a potentially large number of clients with varying computational resources and network speeds presents a practical challenge for real-time deployment in resource-constrained agricultural settings.
    \item The framework performance, particularly the cross-domain results, depends on the quality and diversity of the client data. While the weighted aggregation helps balance contributions, the presence of systematically poor-quality or mislabeled data across multiple clients could still negatively influence the global model, an issue that our current design does not explicitly resolve.
\end{itemize}
\section{Conclusion and future direction}
This study efficiently addressed the critical issue of performance generalization in AI-based plant disease diagnosis. By moving beyond small-scale, homogenous datasets, we introduced the large-scale and globally representative FloraSyntropy Archive, providing an essential benchmark for the agriculture community. The proposed FloraSyntropy-Net framework, integrating a novel Deep Block for enhanced feature learning and a weighted federated optimization scheme, demonstrated superior performance, achieving a SOTA accuracy of 96.38\%. Its exceptional cross-domain adaptation capability, validated by a 99.84\% accuracy on an unrelated insect pest dataset, underscores its robustness and its effectiveness in bridging the generalization gap that has limited previous models. This work provides a comprehensive, privacy-preserving, and scalable solution that is directly applicable to real-world agricultural settings, offering a significant step toward safeguarding global food security through reliable automated diagnosis.
    
    While the results are highly promising, this work opens several avenues for future research. The current framework computational and coordination demands in a federated setting present a practical limitation for real-time use on resource-constrained devices. Our immediate future work will, therefore, focus on developing a lightweight and asynchronous version of the FL protocol to enhance scalability and reduce synchronization overhead. Furthermore, the model performance is inherently tied to the quality of the client data. To address this, we plan to integrate robust data validation and automated quality control mechanisms at the client level to filter out noisy or mislabeled samples before they influence the global model aggregation.

\backmatter

\section*{Declarations}
\bmhead{Ethics approval and consent to participate}
\bmhead{Consent for publication}
Not applicable
\bmhead{Data availability}
Data will be made available on request.
\bmhead{Funding}
No funding
\bmhead{Declaration of competing interest}
The authors declare that they have no known competing financial interests or personal relationships that could have appeared to influence the work reported in this paper.
\bmhead{CRediT authorship contribution statement}
Saif Ur Rehman Khan \& Muhammed Nabeel Asim: Conceptualization, Data curation, Methodology, Software, Validation, Writing original draft \& Formal analysis. Sebastian Vollmer: Conceptualization, Funding acquisition, Supervision. Andreas Dengel: review \& editing. 

\bibliography{sn-bibliography}

\end{document}